\documentclass[lettersize,journal]{IEEEtran}
\usepackage{amsmath,amsfonts}
\usepackage{array}
\usepackage[caption=false,font=normalsize,labelfont=sf,textfont=sf]{subfig}
\usepackage{textcomp}
\usepackage{stfloats}
\usepackage{url}
\usepackage{verbatim}
\usepackage{graphicx}
\usepackage{cite}
\usepackage{xcolor}
\usepackage{multirow}
\usepackage{makecell}
\usepackage[ruled]{algorithm2e}
\usepackage{caption}

\begin{document}

\title{Relational Contrastive Learning and Masked Image Modeling for Scene Text Recognition}

\author{Tiancheng Lin, Jinglei Zhang, Yi Xu, Kai Chen, Rui Zhang, Chang-Wen Chen,~\IEEEmembership{Fellow,~IEEE}
\thanks{This work was supported in part by NSFC 62171282, 111 project BP0719010 and STCSM 22DZ2229005. (\textit{Tiancheng Lin and Jinglei Zhang contributed equally to this research; Corresponding author: Yi Xu.})}
\thanks{Tiancheng Lin, Jinglei Zhang, Yi Xu, Kai Chen, Rui Zhang are with the MoE Key Lab of Artificial Intelligence, AI Institute, Shanghai Jiao Tong University, Shanghai, China. (e-mail:\{ltc19940819, zhangjinglei168, xuyi, kchen, zhang\_rui\}@sjtu.edu.cn)}
\thanks{Chang-Wen Chen is with the Hong Kong Polytechnic University, Hong Kong, China (e-mail: changwen.chen@polyu.edu.hk)}
}

\markboth{Journal of \LaTeX\ Class Files,~Vol.~14, No.~8, August~2021}%
{Shell \MakeLowercase{\textit{et al.}}: A Sample Article Using IEEEtran.cls for IEEE Journals}



\maketitle

\begin{figure*}[t]
  \centering
  
\end{figure*}

\begin{abstract}
Context-aware methods have achieved remarkable advancements in supervised scene text recognition (STR) by leveraging semantic priors from words. Considering the heterogeneity of text and background in STR, we propose that such contextual priors can be reinterpreted as the relations between textual elements, serving as effective self-supervised labels for representation learning.
However, textual relations are restricted to the finite size of the dataset due to lexical dependencies, which causes the problem of over-fitting, thus compromising the quality of representation. 
To address this issue, our work introduces a unified framework of \underline{R}elational \underline{C}ontrastive Learning (RCL) and
\underline{M}asked Image Modeling (MIM) for \underline{STR} (abbreviated as RCMSTR), which explicitly models the \textit{enriched} textual relations.
For the RCL branch, we first introduce the relational rearrangement module to cultivate new relations on the fly. 
Based on this, we further conduct relational contrastive learning to model the intra- and inter-hierarchical relations for frames, sub-words and words.
On the other hand, MIM can naturally boost the context information via masking, where we find that the block masking strategy is more effective for STR.
For the effective integration of RCL and MIM, we also introduce a novel decoupling design aimed at mitigating the impact of masked images on contrastive learning. Additionally, to enhance the compatibility of MIM with CNNs, we propose the adoption of sparse convolutions and directly sharing the weights with dense convolutions in training.
The proposed RCMSTR demonstrates superior performance in various evaluation protocols for different STR-related downstream tasks, outperforming the existing state-of-the-art self-supervised STR techniques.
Ablation studies and qualitative experimental results further validate the effectiveness of our method.
The code and pre-trained models will be available at \url{https://github.com/ThunderVVV/RCMSTR}.
\end{abstract}

\begin{IEEEkeywords}
Scene text recognition, representation learning, contrastive learning,  masked image modeling.
\end{IEEEkeywords}

\section{Introduction}
\label{sec:intro}


\IEEEPARstart{S}{elf} supervised learning (SSL), especially contrastive learning methods~\cite{npid,cpc,he2020momentum,chen2020simple} and masked image modeling~\cite{beit,mae,simmim}, has achieved great success in computer vision tasks for natural images.
For scene text recognition (STR), learning useful visual representation directly from unlabeled data is also attractive since a mass of labeled data is usually needed for training to decode the contained text from images~\cite{fang2021read,str2,str3}.  
However, most of the existing contrastive learning and masked image modeling methods are designed for natural images.
Directly applying them to scene text images is sub-optimal since the characteristics of scene text images are quite different from natural images. 
Specifically, scene text images exhibit the following essential characteristics. 
First, textual foreground and scene background are heterogeneous in scene text images, while text recognition relies primarily on text rather than the background. 
Second, most text images  have only one dominant orientation, $e.g.$, left-to-right reading order for English scripts.
Third, besides the whole word, one text image also contains the sequence of characters and phrase structure of multiple granularities along with the reading order. 
This paper focuses on self-supervised representation learning for STR, seeking to answer the critical question: \textit{Can we exploit these text characteristics to learn effective features from
scene text images without manual annotation?}


\begin{figure*}[t]
    \centering
    \captionsetup{type=figure}
    \includegraphics[width=1\textwidth]{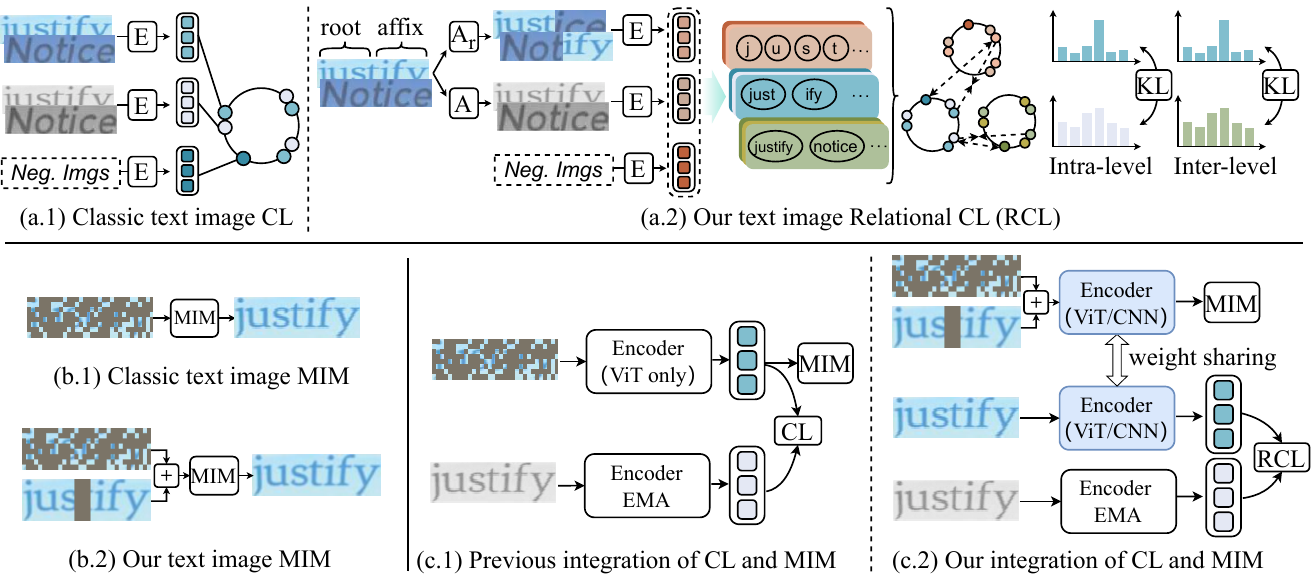}
    \caption{We propose RCMSTR, a unified SSL method for text images to fully utilize textual relations. RCMSTR learns richer textual relations in contrastive learning via rearrangement, hierarchy and interaction. Besides, the MIM simultaneously conducts patch and character-level reconstruction to fully learn the local and global relations of text. Finally, RCMSTR is integrated as a unified SSL framework characterized by a decoupled design and effective compatibility with both CNN and ViT.}
    \vspace{-0.5cm}
    \label{fig:comparison}
    
\end{figure*}


Some pioneering works~\cite{aberdam2021sequence,liu2022perceiving,dig} have explored how to construct variants of contrastive learning (CL) for text recognition. SeqCLR~\cite{aberdam2021sequence} considers scene text images as a sequence of subwords and thus proposes an instance-mapping function, which makes the atoms of CL to be sequential frames ($i.e.$, the subwords) rather than images ($i.e.$, the whole image). PerSec~\cite{liu2022perceiving} conducts CL on low-level and high-level features, aiming to simultaneously learn the representations from stroke and semantic context. 
Masked image modeling (MIM) is another kind of effective SSL approach and some previous works have introduced it to scene text recognition. DiG~\cite{dig} directly combines contrastive learning and masked image modeling into a unified model for text recognition.
However, the gains mainly come from the powerful ViT~\cite{vit}, and the masking strategy is the direct transfer of SimMIM~\cite{simmim}.
In the same period, MaskOCR~\cite{maskocr} proposed to use vertical patches as ViT inputs for MIM, while the vertical patches are too coarse to model detailed stroke relations. 

These above SSL methods are mainly transferred from natural images and only partially explore the text characteristics. 
Unlike them, our idea is rooted in the supervised text recognition in our community, aiming at fully exploring the characteristics of the text. 
In particular, context-aware methods~\cite{context1,context2,fang2021read} achieved great success by incorporating semantic priors from words in a supervised fashion. 
We argue that such contextual information can be interpreted as the relations of textual primitives and thus can be utilized in an unsupervised way. 
This insight motivates the proposed unified framework, named RCMSTR, of contrastive learning and masked image modeling for scene text images.
The key challenge here is that textual relations are restricted to the finite size of the dataset, which usually causes the problem of over-fitting due to lexical dependencies~\cite{wan2020vocabulary}. 
To address this problem, we equip RCMSTR with several novel modules, where the overall comparison of  existing methods is shown in Fig.~\ref{fig:comparison}. 
Different from existing CL methods, our text image CL is based on relational CL (RCL)~\cite{zheng2021ressl} to model the  intra- and inter-hierarchical relations simultaneously with enriched textual relations. To achieve this, we first create permuted images  with diverse relations on the fly by rearranging images. 
Then, we explicitly model the intra-hierarchical relations of primitives in different granularities, ranging from the words, roots, affixes  and characters, to learn semantic information and enhance representation learning.
We further propose consistency constraints to explore the inter-hierarchical relations considering their similar attributes in color and stroke.
Meanwhile, different from existing MIM methods, we incorporate a block masking strategy into the widely used random masking to learn richer relations.
Finally, RCMSTR effectively integrates RCL and MIM through a decoupling design, demonstrating compatibility with both ViT and CNN encoders. By decoupling  the MIM encoder from CL, the feature mismatch problem of masked and non-masked images is addressed. RCMSTR further utilizes sparse convolution, enhancing its compatibility with CNN architectures.

We conduct self-supervised learning on SynthText~\cite{gupta2016synthetic} for both CNN and ViT, which are evaluated on downstream tasks, $e.g.$, STR representation quality, supervised fine-tuning, semi-supervised fine-tuning and semantic segmentation. 
RCMSTR outperforms all compared SSL algorithms by a large margin on STR representation quality for both CNN and ViT. RCMSTR significantly outperforms SeqMoCo baseline on STR supervised fine-tuning and semi-supervised fine-tuning. Also, it achieves large performance gains on TextSeg~\cite{textseg} for semantic segmentation. Moreover, extensive ablation experiments are conducted to verify our key model components.

We first presented \underline{R}elational \underline{C}ontrastive \underline{L}earning for \underline{S}cene \underline{T}ext \underline{R}ecognition
(RCLSTR)~\cite{rclstr} published in ACMMM-2023. 
This paper is a
significant extension with the following four improvements:

\begin{itemize}
    \item We present a novel framework featuring new notations, formulas, and diagrams, complemented by comprehensive methodology descriptions and associated learning objectives.
    \item We explore the MIM on text images and propose a novel masking strategy with both patch and character-level masking. This innovative MIM design encourages the network to fill in masked characters to promote the learning of global context relations.
    \item We design a unified RCLSTR-MIM integrated framework with an effective decoupling design to achieve higher performance. Furthermore, we overcome the ViT-only limitations by leveraging sparse convolution, ensuring thorough compatibility with both CNN and ViT architectures.
    \item We evaluate RCMSTR on more STR datasets (12 in total) and more downstream tasks ($e.g.$, text segmentation), showing that RCMSTR surpasses RCLSTR by a large margin, as well as other previous SSL methods. Besides, we conduct our experiments on both CNN and ViT, demonstrating the flexibility of RCMSTR.
        
    
\end{itemize}

\section{Related Work}

\subsection{Self-Supervised Learning on natural images}
For natural image self-supervised learning, contrastive learning methods~\cite{npid,cpc,he2020momentum,chen2020simple,swav,byol,simsiam,dino} and masked image modeling methods~\cite{mae,beit,simmim,ibot,dinov2} show great success, and they conduct discrimination-based and reconstruction-based pretext task for pre-training, respectively.

Contrastive learning methods perform the instance discrimination task to classify different data-augmentation views from the same image into a class. In NPID~\cite{npid}, the task of instance discrimination is proposed, and noise contrastive estimation (NCE) is used for contrastive learning, which is further replaced by InfoNCE~\cite{cpc}. MoCo~\cite{he2020momentum} and SimCLR~\cite{chen2020simple} improve the quality of learned representations, which propose the momentum encoder and use a single network with a large batch size, respectively. SwAV~\cite{swav} constrains the consistency of cluster allocation of different data-augmentation views. Some subsequent studies, such as BYOL~\cite{byol}, SimSiam~\cite{simsiam}, and DINO~\cite{dino}, propose asymmetric frameworks that eliminate the need for negative samples.
On the other hand, some works~\cite{zheng2021ressl,wang2022contrastive} propose to use KL divergence to constrain relative consistency in the form of similarity distribution.

Masked image modeling methods randomly mask input image patches and perform the reconstruction task. In BEiT~\cite{beit}, the visual token reconstruction task is proposed, and they utilize a tokenizer to obtain discrete visual tokens as reconstruction targets. MAE~\cite{mae} and SimMIM~\cite{simmim} further replace the reconstruction target from tokens to pixels, overcoming the need for an additional discrete tokenizer. More recently, iBOT~\cite{ibot} and DINOv2~\cite{dinov2} used the momentum encoder as an online tokenizer. 


However, these self-supervised methods are designed for natural images, which is quite different from text images. Considering the characteristics of text, we need to specially design the self-supervised method for text images.

\subsection{Self-Supervised Text Recognition}

Some pioneering works~\cite{whatif,aberdam2021sequence,liu2022perceiving,dig,siman,guan2023self} explored self-supervised methods in text recognition and have achieved promising results. We summarize these methods into three main categories. The first approach is based on contrastive learning. SeqCLR~\cite{aberdam2021sequence} maps sequence features of words to instances as atomic elements of contrast learning. They only consider the text sequence structure. PerSec~\cite{liu2022perceiving} proposed to conduct contrastive learning on the low-level stroke and the high-level semantic features of text images corresponding to visual and semantic information. The second is based on masked image modeling. DiG~\cite{dig} proposed a self-supervised framework for text
recognition that combines contrastive learning and masked image models(MIM). Concurrent work~\cite{maskocr} also uses MIM for STR. However, these MIM methods are directly transferred from natural image methods. The third is based on generative learning, and a representative work is  SimAN~\cite{siman}, which proposes to reconstruct the images from the decoupled content and style information.
In sum, the above methods have not fully explored the characteristics of text images. Our approach takes into account the heterogeneity of texts, the left-to-right structure and the hierarchical structure of the sequence to fully explore the text characteristics. We propose a novel contrastive learning framework to enrich the textual relations via rearrangement, hierarchy and interaction.

\subsection{Text Recognition}

Recent advancements in text recognition mainly follow the sequence-to-sequence paradigm, known for its flexibility and convenience derived from an end-to-end approach. Text recognition methods based on sequence-to-sequence models can be divided into three categories according to the decoder: CTC-based decoder, attention-based decoder, and transformer-based decoder. 
The majority of CTC-based text recognition methods~\cite{he2016reading, su2017accurate} follow similar pipelines. The representative CRNN~\cite{shi2016end} algorithm is comprised of a CNN feature extractor, an RNN sequence encoder and a CTC decoder. 
On the other front, Shi et al.~\cite{shi2018aster} introduced a text recognition model that employs an attention decoder alongside a spatial transform network to address the challenges of handling irregular text recognition. Spatial attention decoders have been further incorporated by Yang et al.~\cite{yang2017learning} and Wojna et al.~\cite{wojna2017attention}, demonstrating their effectiveness in enhancing text recognition capabilities. In the recent landscape, spurred by the rapid ascent of Transformers, some text recognition methods~\cite{lee2020recognizing, lu2021master} have embraced Transformer decoders in their models.

\section{Method}

\begin{figure*}[t]
\normalsize
  \centering
  \includegraphics[width=1\textwidth]{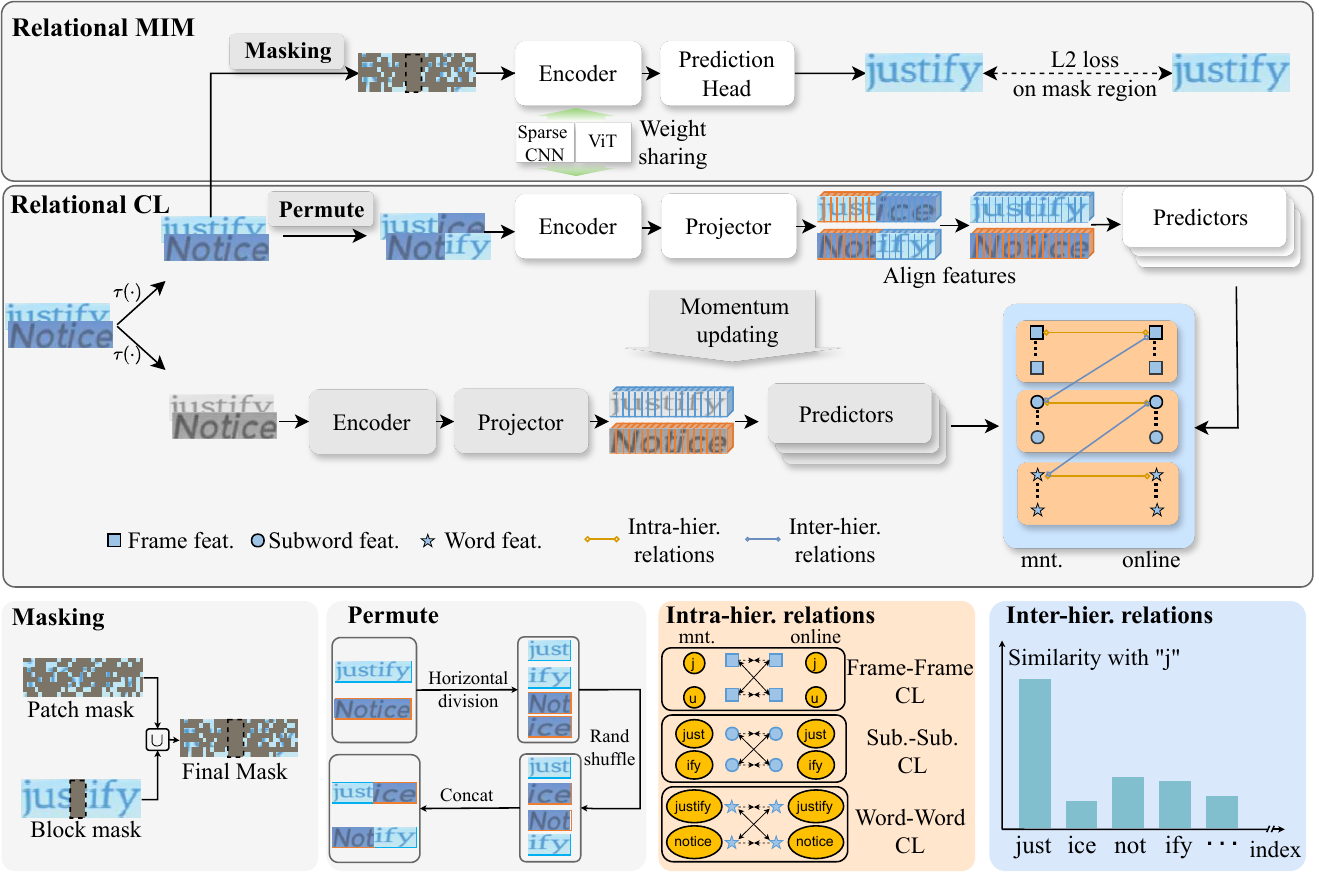}
  \caption{\textbf{Block diagram.} Each image in a batch is augmented and processed by the Relational MIM and CL components. In MIM, the image undergoes masking based on a specific strategy to facilitate local and global relational modeling, followed by a prediction head that reconstructs the masked regions. In CL, the image is augmented twice and then fed separately into the online branch (top) and the momentum branch (bottom) of the encoder and projector to create pairs of representation maps. In the module of enriching relations, we randomly permute the image patches and reverse the permutation on their features. Next, for the hierarchical contrastive learning of these representations, we apply three predictors that transform them into frames, subwords and words, respectively. Finally, we apply the relational contrastive loss on the corresponding intra- and inter-hierarchical positive pairs.}
  
  \label{fig:rcmstr}
\end{figure*}

\subsection{Overall Framework}

We proposed a self-supervised framework for scene text recognition with two branches of relational contrastive learning and relational masked image modeling, as shown in Figure~\ref{fig:rcmstr}. First, 
the input image is augmented to obtain two views (including the first augmented view and the second augmented view), which are input into the online and momentum encoder, respectively, to generate two sets of features and perform relational contrastive loss. On the other hand, the first augmented view is fed into a decoupled masked image modeling branch, with a masking strategy for local and global relational modeling of the image. We use the online encoder and a prediction head to reconstruct the masked images. 
Finally, the online encoder is updated by back-propagation supervised by the summation of relational contrastive loss and reconstruction loss, while the momentum encoder is a smoothly evolved version of the online encoder.

\subsection{Relational Contrastive Learning}
\noindent\textbf{Motivation:} 
Due to the finite size of the dataset, the textual relations are restricted, and the performance of naive relational contrastive learning is limited. Therefore, we propose to learn the intra- and inter-hierarchical relations with the enriched textual information, building a more complete relational contrastive learning framework.
1) Text images can be divided and rearranged into new context relations. For example, as shown in Figure~\ref{fig:comparison} (a.2), words ($e.g.$, ``justify" and ``notice") can be broken up into subwords of roots and affixes and new words of ``justice" and  ``notify'' can be achieved by rearrangement (Sect.~\ref{sec:regularization}). 
2) There are multi-level relations in text images, such as words, subwords and characters (Sect.~\ref{sec:hierarchical}). 
3) We can leverage the interactions among the objects of different levels. For example, the characters (at the lowest level) and subwords (at the middle level) from the same locations (in the same images) share similar attributes in color and stroke, thus showing higher similarity in the feature space (Sect.~\ref{sec:consistency}). 

\noindent\textbf{Preliminary:} 
Based on the structure of MoCo~\cite{he2020momentum}, an efficient and effective baseline, we propose the relational momentum contrastive learning framework for the CL branch.
The pretext task, instance~\footnote{
Since STR models usually encode a text image as a sequence of features, we will discuss instances at different levels in sect.~\ref{sec:hierarchical}.} discrimination~\cite{npid},
aims to train the model to discriminate the positive view from the negative views. 
The query view $\mathrm{X}_i^q$ and positive view $\mathrm{X}_i^p$ are encoded as $\mathbf{q}$ and $\mathbf{p}$. To avoid the need for large batch size, we follow MoCo to maintain a queue of size $K$, and there are $K$ negative features $\{ \mathbf{n}_k\}_{k=1}^K$ from other images. Then, the contrastive loss of InfoNCE is written as:
\begin{equation}
\resizebox{0.9\hsize}{!}{$\mathcal{L}_{info}(\mathbf{q}, \mathbf{p}, \mathbf{n}) = -\log \frac{\exp (\mathbf{q}\cdot \mathbf{p} / \tau_{info})}{\sum_{\mathbf{u} \in \{ \mathbf{n}_k\}_{k=1}^K \cup \{\mathbf{p} \}} \exp(\mathbf{q}\cdot \mathbf{u} / \tau_{info})}\,, $}
\label{eq:infoloss}
\end{equation}
where $\tau_{info}$ is a temperature hyperparameter. This loss function aims to pull features of positive pairs closer together and push all the other negative examples farther apart.

Naive relational contrastive learning~\cite{zheng2021ressl,wang2022contrastive}
aims at learning not only the relation between query views and positive views but also the relation between query views and negative views. 
To achieve this, 
we calculate the similarity between the positive and the negatives ($i.e.$, $P$) and that between the query and the negatives ($i.e.$, $Q$), and encourage the agreement of two similarity distributions. Formally, we use symmetric Kullback-Leibler (KL) Divergence as the measure of disagreement, imposing consistency between $P$ and $Q$:

\begin{align}
\begin{split}
Q_i(\mathbf{q}, \mathbf{n}) & = \frac{\exp(\mathbf{q} \cdot \mathbf{n}_i / \tau_{kl})}{\sum_{k=1}^K \exp(\mathbf{q} \cdot \mathbf{n}_k / \tau_{kl})}, \\
P_i(\mathbf{p}, \mathbf{n})  & = \frac{\exp(\mathbf{p} \cdot \mathbf{n}_i / \tau_{kl})}{\sum_{k=1}^K \exp(\mathbf{p} \cdot \mathbf{n}_k / \tau_{kl})}, \\
\mathcal{L}_{kl}(\mathbf{q}, \mathbf{p}, \mathbf{n})  & = \frac{1}{2} D_{\mathrm{KL}} (P \Vert Q) + \frac{1}{2} D_{\mathrm{KL}} (Q \Vert P),
\end{split}
\label{eq:klloss}
\end{align}
where $\tau_{kl}$ is also a temperature hyper-parameter. The total relational loss is a weighted average of the InfoNCE loss term and the KL loss term: 
\begin{equation}
\mathcal{L}_{re}(\mathbf{q}, \mathbf{p}, \mathbf{n}) = \mathcal{L}_{info}(\mathbf{q}, \mathbf{p}, \mathbf{n}) + \alpha \mathcal{L}_{kl}(\mathbf{q}, \mathbf{p}, \mathbf{n}), \label{eq:relationloss}
\end{equation}
where $\alpha$ denotes the coefficient to balance two terms. $\alpha$ is empirically set to 0.3. 
The first term is the absolute similarity constraint between $\textbf{q}$ and $\textbf{p}$. The second term is the relative similarity constraint, which aims to keep the similarity distribution consistency of $\textbf{q}$ and $\textbf{p}$ with the negatives.



\subsubsection{\textbf{Enriching textual relations via permutation operation}}
\label{sec:regularization}

Previous works~\cite{wan2020vocabulary,zhang2022context} have pointed out that text recognizers are prone to over-dependence on context in supervised learning, while our method advances them further in unsupervised learning. 
To alleviate this context-dependent problem, we propose a permutation module to generate new text images. The generated images contain more diversity of context relations, encouraging the encoder not to over-fit finite contexts in the dataset.
The default process goes like 1) uniformly dividing text images into multiple horizontal patches, 2) randomly shuffling and concatenating patches to generate permuted images, and 3) adding a regularization loss term corresponding to these permuted images after alignment. For the step of image division, we explore different variant operations in ablation studies.


Specifically, the permutation operation is performed directly on the input images, as shown in Figure~\ref{fig:rcmstr}. Firstly, we divide each image horizontally into $N$ patches, where the default $N$ is 2. Next, we take $M$ images as a group to randomly shuffle the $NM$ patches in each group, where the default $M$ is 2. Then, every $N$ patches are concatenated horizontally to achieve new images with the enriched textual information, denoted as $\mathbf{x}^{e}$. 
Note that $\mathbf{x}^{e}$  is only fed into the online encoder and projector for features, and such an implementation empirically performs better, likewise the multi-cropping strategy~\cite{swav}. To align all features of permuted images, we unshuffle them (inverting the random shuffle operation) to put the features back in their original position. We denote the resulting features of regularization as $\mathbf{q}^{e}$. The relational contrastive loss with the enriched images can be written as:
\begin{equation}
\mathcal{L}_{ere}(\mathbf{q}, \mathbf{p}, \mathbf{n}) = \frac{1}{2}\mathcal{L}_{re}(\mathbf{q}, \mathbf{p}, \mathbf{n})  + \frac{1}{2}\mathcal{L}_{re}(\mathbf{q}^{e}, \mathbf{p}, \mathbf{n}),\label{eq:regularization}
\end{equation}
where $\mathcal{L}_{re}$ is from Equation~\ref{eq:relationloss}. $\mathcal{L}_{ere}$ constrains the invariance of the relation under random permutations. 



\subsubsection{\textbf{Modeling intra-hierarchical relations}}
\label{sec:hierarchical}

Given that text images are encoded as  sequence features, contrastive learning can be implemented across various combinations of elemental features within these sequences. Considering that text words have different granularities in the horizontal direction, we propose a novel hierarchical structure that maps the features to three levels: frame, subword, and word. Thus, we conduct hierarchical relational contrastive learning to learn about relations at each level. 

To achieve this, we employ three pairs of predictors using average pooling functions with different horizontal window sizes. For frame-level character objects, we use the average pooling function with window size 1 to project encoder outputs to frame-level features. At the subword level, addressing elements like roots and affixes, we average projected features into $T$ subwords ($T=4$ by default) since subwords generally comprise multiple letters. Finally, the word-level representation considers entire words, applying an averaging function to the full feature sequences.
At each level, we maintain a separate queue of negative features, respectively. We calculate the relational contrastive losses at each level and sum them up:
\begin{equation}
\mathcal{L}_{hier} =  \sum_{h \in H}\mathcal{L}_{ere}(\mathbf{q}^h, \mathbf{p}^h, \mathbf{n}^h),
\end{equation}
where $H=\{frame, subword, word\}$. With the proposed hierarchical relational contrastive learning, the model can learn the frame-frame, subword-subword and word-word relations simultaneously.

\subsubsection{\textbf{Modeling inter-hierarchical relations}}
\label{sec:consistency}

Besides the relation within each level, there are semantic relations between features across different hierarchies. Therefore, we propose consistency constraints to learn the relation between neighboring levels. Our default implementation performs frame-subword and subword-word consistency constraints, and we provide the results of other settings in the experiments. As shown in Figure~\ref{fig:rcmstr}, for frame-subword relations, since the frame and subword features from the same spatial locations (in the same images) show higher similarity in the feature space, we treat them as positives and treat features in other locations as negatives. The subword-word positives and negatives are determined in the same way.
For the frame-subword and subword-word relation, we impose the consistency between the similarity distributions by KL loss as the measure of disagreement:
\begin{align}
\begin{split}
\mathcal{L}_{f2s} & = \mathcal{L}_{kl}(\mathbf{q}^{f}, \mathbf{p}^{s}, \mathbf{n}^{s}), \\
\mathcal{L}_{s2w} & = \mathcal{L}_{kl}(\mathbf{q}^{s}, \mathbf{p}^{w}, \mathbf{n}^{w}),
\end{split}
\end{align}
where superscripts of $\{f$, $s$, $w\}$ denote $\{frame$, $subword$, $word\}$, respectively. This loss constrains the relation between each feature and its neighboring upper-level feature. Here, we take the positives and negatives from the upper level because we consider that upper-level features contain more semantic information than those at lower-levels.
Finally, the total loss of our relational contrastive learning is formulated as:
\begin{equation}
\resizebox{0.9\hsize}{!}{$\mathcal{L}_{RCL} = \underbrace{\sum_{h \in H}\mathcal{L}_{ere}(\mathbf{q}^h, \mathbf{p}^h, \mathbf{n}^h)}_{\text{Intra-hierarchical relations}} + \underbrace{\vphantom{\sum_{h \in H}L_{ere}(\mathbf{q}^h, \mathbf{p}^h, \mathbf{n}^h)}\mathcal{L}_{f2s} + \mathcal{L}_{s2w}}_{\text{Inter-hierarchical relations}}$}
\end{equation}

\subsection{Relational Masked Image Modeling}
\noindent\textbf{Motivation:} Contrastive learning enables feature representation with better discrimination but fails to capture the structure and texture information in spatial dimensions~\cite{CLMIM}. However, the low-level strokes, $e.g.$, font, color and style of texts, are also necessary for STR~\cite{liu2022perceiving}. 
Therefore, we propose the relational MIM framework for text recognition.
In Sect.~\ref{sec:mim1}, we propose a relational masking strategy to fully model the relations based on local and global reconstruction.
In Sect.~\ref{sec:mim2}, we design a decoupling framework to effectively integrate the capabilities of relation CL and MIM. Meanwhile, we propose a carefully designed framework that integrates powerful MIM capabilities for CNN architectures in Sect.~\ref{sec:mim3}. 

\subsubsection{\textbf{Relational Masking Strategy}}
\label{sec:mim1}


Scene text images  typically exhibit two levels of structure: the local stroke structure and the global sequential characters structure. Previous text MIM methods~\cite{dig} did not consider both levels simultaneously. Thereby, the relations cannot be fully captured. To overcome this limitation, we propose to simultaneously conduct patch-level and character-level reconstruction to comprehensively model the local and global relationships in text. 


\begin{figure}[t]
\centering
\includegraphics[width=1\columnwidth]{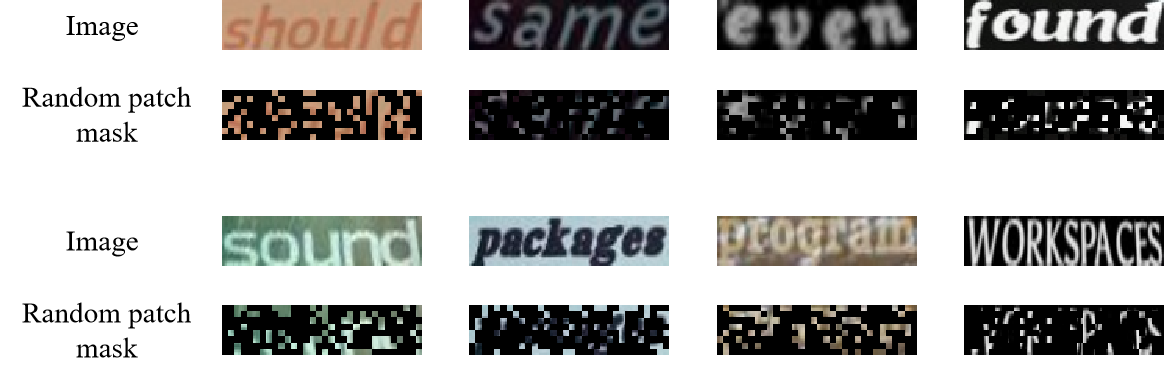}
\caption{Random masking of patches.}
\label{fig:mim_local}
\end{figure}

For the modeling of local relations of text images, a random masking strategy of image patch alignment is adopted, as shown in Figure~\ref{fig:mim_local}. Given that the image patch is the fundamental processing unit for ViT, applying masks at the patch level is both straightforward and effective. Each image patch is either entirely visible or entirely masked. 

\begin{figure}[t]
\centering
\includegraphics[width=1\columnwidth]{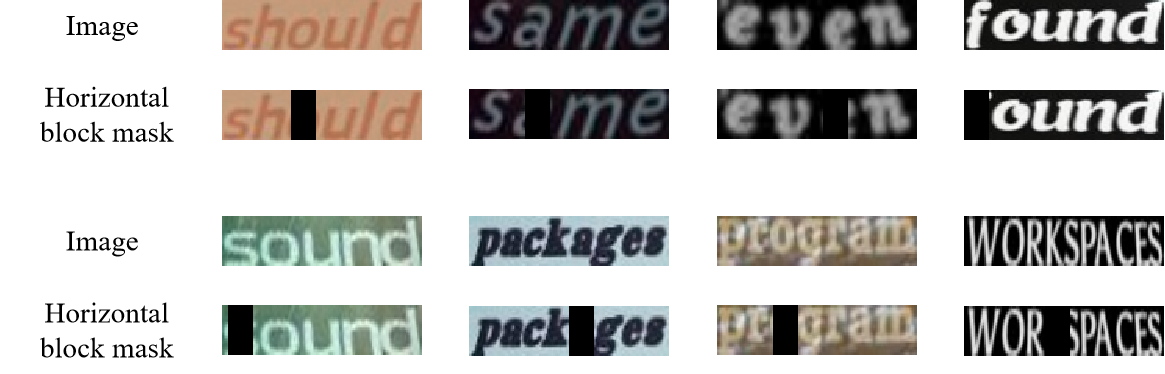}
\caption{Horizontal block masking.}
\label{fig:mim_global}
\end{figure}

For the modeling of the global relations of text images, we propose the block-masking strategy in the horizontal direction, as depicted in Figure~\ref{fig:mim_global}, where a fixed pixel width is used to define a horizontal block. 
We randomly select blocks in the horizontal direction of the image to mask, setting the default number of masked blocks to 1. In the ablation experiment, we study the performance of different numbers of block-masking.

\textbf{Reconstruction loss:} We employ the normalized RGB values of the original pixels as the targets for reconstruction. Each pixel value in the original image is initially normalized to fall within the $[0, 1]$ range, serving as the label for reconstruction. MSE loss, $i.e.$, $L_2$ loss, is used for constraint:

\begin{equation}
    \mathcal{L}_{MIM} = \frac{1}{3N} \sum_{i \in N}\sum_{j \in \{R,G,B\}}{(x_{ij} - y_{ij})^2}, 
\end{equation}
where $x_{ij}$ and $y_{ij}$ are the predicted pixel value and the original pixel value, respectively, and $N$ is the number of masked pixels.

\subsubsection{\textbf{Decoupling Design}}
\label{sec:mim2}

Contrastive learning and MIM are utilized for self-supervised learning on text images for different purposes. However, effectively integrating the capabilities of these two methodologies poses a challenge. We found that directly coupling both methods~\footnote{The coupling design denotes feeding masked images into the CL online encoder. The resulting feature would be utilized for both MIM reconstruction and serving as positive features for CL. The diagram of this coupling design can be referenced in the DiG~\cite{dig} paper.} as in the previous DiG~\cite{dig} may lead to unstable performance gains. Similar instability issues have also been reported in previous research~\cite{ibot}, attributed to the influence of masked and non-masked positive image pairs. We hypothesize that it may arise from the distribution mismatch of masked and non-masked images in contrastive learning. To alleviate this instability problem, we propose a decoupling design that involves separately feeding non-masked and masked images into contrastive learning and MIM, respectively. The decoupling design ensures that the contrastive learning part does not receive masked images as input, thereby mitigating instability issues.

The meaning of decoupling is that masked images are only used for reconstruction, while contrastive learning only utilizes unmasked images. Specifically, as illustrated in Figure~\ref{fig:rcmstr}, the input image undergoes data augmentation, resulting in three augmented views. The first augmented view undergoes image masking and is then individually fed into the MIM component. The other two augmented views, without image masking, are utilized as positive pairs and fed into two branches of contrastive learning. At the same time, weight sharing is employed between the MIM encoder and the online encoder of contrastive learning. Consequently, our framework reduces the interference between the two components and enables the encoders to derive benefits from the learning gains of both.  We provide pseudocode in supplementary materials.

The final loss is the weighted summation of the relational contrastive loss $L_{RCL}$ and masked image modeling loss $L_{MIM}$. It is denoted as

\begin{equation}
    \mathcal{L} = \mathcal{L}_{MIM} + \beta \mathcal{L}_{RCL}, 
\end{equation}
where $L_{RCL}$ and $L_{MIM}$ represent loss functions for RCL and MIM, respectively. $\beta$ is a weight factor, empirically set to 0.1.

\subsubsection{\textbf{CL-MIM Integration for CNNs}}
\label{sec:mim3}

Previous text recognition CL-MIM integrated methods~\cite{dig} were exclusively designed for ViT, but they are unsuitable for CNNs. Despite the performance advantages of ViT-based text recognition methods over CNN, many text recognition applications still employ CNN-based methods due to their computational and deployment efficiency. Therefore, we aspire to incorporate the capability of MIM into the architectures of CNNs. However, integrating MIM into CNN methods for text recognition encounters two main challenges. 
The primary challenge is how to adapt CNN to ensure the effectiveness of MIM, as a study by L. Jing et al.~\cite{maskedsiamese} suggests that directly masking images is ineffective. The second challenge lies in effectively integrating contrastive learning and MIM in a unified CNN framework. 

Regarding the solution to these challenges, our key idea is to replace all standard convolutions with sparse convolutions, which is inspired by ConvNextv2~\cite{convnextv2}. ConvNextv2 discovered that MIM pre-training with sparse convolutions is effective for \textit{fine-tuning} with standard convolution without requiring additional handling. 
Inspired by this idea, we propose to use sparse convolution to realize MIM pre-training --- the masked inputs are treated as sparse patches processed by the sparse convolutions. However, the contrastive learning branch originally used standard convolutions, which is incompatible. We found that sparse convolution is actually equivalent to standard convolution for the original, unmasked inputs. So, we also use sparse convolutions in the CL branch. Therefore, both branches can be implemented in a unified way. This approach enables the adaptability of MIM on CNNs and effectively combines the capabilities of both.

\section{Experiments}

         



\subsection{Implementation Details}

\noindent\textbf{Datasets.} We conduct our experiments on public datasets of scene text recognition. For our self-supervised framework pre-training, we follow SeqCLR~\cite{aberdam2021sequence} to use the synthetic dataset \textbf{SynthText}~\cite{gupta2016synthetic}. For evaluation, we use twelve real-scene text datasets, including IIIT5K-Words (\textbf{IIIT5K})~\cite{IIIT5K}, ICDAR 2003 (\textbf{IC03})~\cite{ic03}, ICDAR 2013 (\textbf{IC13})~\cite{ic13}, Street View Text (\textbf{SVT})~\cite{svt}, ICDAR 2015 (\textbf{IC15})~\cite{ic15}, Street View Text Perspective (\textbf{SVTP})~\cite{svtp}, \textbf{CUTE80}~\cite{cute}, COCOText-Validation (\textbf{COCO})~\cite{cocotext}, \textbf{CTW} dataset~\cite{ctw}, Total-Text dataset (\textbf{TT})~\cite{tttext}, weakly occluded scene text (\textbf{WOST})~\cite{ost} and heavily occluded scene text (\textbf{HOST})~\cite{ost}. They covered regular, perspective and occluded text images.

\noindent\textbf{Metrics.} To evaluate the STR performance, we adopt the metrics of word-level accuracy (Acc), which calculates the number of correctly predicted words divided by the total number of words. 
For the text segmentation task, we assess the  performance using the average intersection over union (IoU).

\noindent\textbf{RCL Network Configurations.} For the RCL branch, we get two data augmentation views and feed them into two separate branches of encoder and head for contrastive learning. 

\begin{itemize}

\item \textbf{Data Augmentation:} For the data augmentation, we follow SeqCLR~\cite{aberdam2021sequence}, and our augmentation consists of a random subset of linear contrast, blur, sharpen, crop, perspective transform and piecewise affine operations. 
We further apply the masking strategy to the first view for MIM, which will be explained in the RMIM network configuration. 

\item \textbf{Encoder:} 
When using CNN architectures, we take blocks of transformation (TPS, Thin Plate Spine)~\cite{shi2016robust} and CNN feature extraction as the encoder, replacing all standard convolutions with sparse convolutions. 
When using the ViT encoder, We use a vanilla ViT~\cite{vit} as the feature encoder with a patch size of $4 \times 4$.

\item \textbf{Projector:} When using CNN architectures, we follow the SeqCLR~\cite{aberdam2021sequence} setting to use a two-layer Bidirectional-LSTM (BiLSTM) with 256 hidden units as the projector. 
When using ViT architectures, we follow DiG~\cite{dig}, and the projector is a 3-layer MLP. In MLP, each layer has a fully connected layer and a normalized layer. 

\item \textbf{Predictor:} 
We set three predictors for the frame, subword, and word levels, each with its own window size. For each level, we use an adaptive averaging layer with the corresponding window size, followed by a final FC layer. When using ViT architectures, we follow DiG~\cite{dig}, and the last layer is a 3-layer MLP.

\end{itemize}

\noindent\textbf{RMIM Network Configurations.} 
For the RMIM branch, we feed the masked images into the feature encoder and use an MLP layer as the prediction head. 

\begin{itemize}

\item \textbf{Masking:} When using CNN architectures, we follow the CNN masking method in ConvNextV2~\cite{convnextv2}. As the convolutional model has a hierarchical design, the mask is generated and upsampled for all stages. When using ViT architectures, our masking strategy consists of small random patches and horizontal blocks, which are superimposed to obtain the final image mask.

\item \textbf{Encoder}: When using CNN architectures, we use the same CNN network as in the RCL but replace all standard convolutions with sparse convolutions. When using ViT architectures, we use the same ViT~\cite{vit} as in the RCL. We follow the typical self-supervised method SimMIM~\cite{simmim} to replace each masked image block with a shared learnable mask embedding.  

\item \textbf{Prediction head}: When using CNN architectures, we follow ConvNextV2~\cite{convnextv2} and use a ConvNext block as the prediction head. It consists of three convolutions with a norm and activation layer. When using ViT architectures, we follow DiG~\cite{dig}, and use the MLP layer as the prediction head. 

\end{itemize}

\subsection{Representation Quality Evaluation.}

\begin{table*}[t]
\setlength{\tabcolsep}{3pt}
\caption{\textbf{Representation quality.} Accuracy(\%) is used to evaluate the quality of representation from encoder, and we train a decoder with labeled data on top of frozen encoder which was pre-trained on unlabeled images. RCLSTR results with different modules added are listed, ``ERE" denotes the module of enriching relations, ``INTRA" denotes the intra-hierarchical relation module and ``INTER" denotes the inter-hierarchical relation module. We further integrate relational MIM on RCLSTR, constructing a unified SSL framework RCMSTR. $^{\text{\textdagger}}$ denotes that the results are reproduced using our datasets.}
\vspace{-0.3cm}
\begin{center}
    \centering
    \begin{tabular}{ccllccccccccccccc}
    \hline
        \multirow{2}{*}{Arch.} & \multirow{2}{*}{Dec.} & \multirow{2}{*}{Method} & & \multicolumn{13}{c}{Scene-Text Dataset}\\ 
        \cline{5-17}
        & & & & IIIT5K & IC03 & IC13 & SVT & IC15 & SVTP & CUTE & COCO & CTW & TT & HOST & WOST & Avg \\ \hline
        \multirow{14}{*}{CNN} & \multirow{7}{*}{CTC} & \multicolumn{2}{l}{SeqCLR~\cite{aberdam2021sequence}} & 35.70 & 43.60 & 43.50 & - & - & - & - & - & - & - & - & - & - \\ 
        ~ & ~ & \multicolumn{2}{l}{PerSec-CNN~\cite{liu2022perceiving}} & 37.90 & 45.70 & 46.40 & - & - & - & - & - & - & - & - & - & -\\ 
        ~ & ~ & \multicolumn{2}{l}{SeqMoCo w/o KL} & 41.63 & 48.21 & 46.50 & 25.35 & 22.05 & 19.53 & 22.22 & 10.00 & 17.01 & 19.15 & 8.94 & 12.58 & 24.43 \\
        ~ & ~ & \multicolumn{2}{l}{SeqMoCo} & 42.97 & 51.44 & 48.37 & 25.35 & 23.01 & 20.62 & 23.26 & 10.69 & 16.24 & 21.01 & 8.77 & 12.87 & 25.38 \\ 
        \cline{3-17}
        ~ & ~ & \multirow{3}{*}{RCLSTR} & w/ ERE & 48.43 & 58.94 & 54.98 & 35.09 & 26.43 & 26.82 & 29.17 & 13.56 & 21.66 & 25.38 & 11.80 & 17.30 & 30.80 \\ 
        ~ & ~ & & w/ ERE-INTRA & 51.90 & 61.36 & 59.01 & 38.79 & 30.62 & 30.08 & 30.21 & 15.39 & 24.84 & 29.47 & 13.25 & 18.87 & 33.65 \\ 
        ~ & ~ & & w/ ERE-INTRA-INTER & \underline{54.83}& \underline{64.82} & \underline{60.89} & \underline{41.58} & \underline{32.60} & \underline{34.26} & \underline{32.64} & \underline{17.36} & \underline{26.62} & \underline{30.92} & \underline{14.53} & \underline{20.94} & \underline{36.00} \\ 
        ~ & ~ & \multicolumn{2}{l}{RCMSTR} & \textbf{64.97} & \textbf{79.01} & \textbf{72.81} & \textbf{59.35} & \textbf{41.65} & \textbf{47.91} & \textbf{44.79} & \textbf{23.31} & \textbf{38.47} & \textbf{39.75} & \textbf{22.10} & \textbf{31.75} & \textbf{47.16} \\ \cline{2-17}
        ~ & \multirow{7}{*}{Attn} & \multicolumn{2}{l}{SeqCLR~\cite{aberdam2021sequence}} & 49.20 & 63.90 & 59.30 & - & - & - & - & - & - & - & - & - & -\\ 
        ~ & ~ & \multicolumn{2}{l}{PerSec-CNN~\cite{liu2022perceiving}} & 50.70 & 65.70 & 61.10 & - & - & - & - & - & - & - & - & - & - \\ 
        ~ & ~ & \multicolumn{2}{l}{SeqMoCo w/o KL} & 50.97 & 58.36 & 55.86 & 35.55 & 29.42 & 28.53 & 30.56 & 14.66 & 21.59 & 25.33 & 13.91 & 18.92 & 31.97 \\
        ~ & ~ & \multicolumn{2}{l}{SeqMoCo} & 51.83 & 59.75 & 59.90 & 37.40 & 31.73 & 28.99 & 32.29 & 16.17 & 22.04 & 26.74 & 13.62 & 21.11 & 33.46 \\ 
        \cline{3-17}
        ~ & ~ & \multirow{3}{*}{RCLSTR} & w/ ERE & 56.30 & 67.70 & 63.25 & 41.27 & 35.05 & 36.90 & 37.15 & 18.21 & 26.88 & 30.24 & 18.13 & 24.46 & 37.96 \\ 
        ~ & ~ & & w/ ERE-INTRA & 59.03 & 71.51 & 67.29 & 46.37 & 38.32 & 36.90 & 36.81 & 20.00 & 30.32 & 33.92 & 20.24 & 26.78 & 40.62 \\ 
        ~ & ~ & & w/ ERE-INTRA-INTER & \underline{61.07} & \underline{72.90} & \underline{68.77} & \underline{50.54} & \underline{40.25} & \underline{40.16} & \underline{39.24} & \underline{22.45} & \underline{32.10} & \underline{35.97} & \underline{21.81} & \underline{28.27} & \underline{42.79} \\ 
        ~ & ~ & \multicolumn{2}{l}{RCMSTR} & \textbf{70.00} & \textbf{83.85} & \textbf{79.80} & \textbf{66.00} & \textbf{49.30} & \textbf{55.19} & \textbf{48.61} & \textbf{27.93} & \textbf{43.95} & \textbf{44.97} & \textbf{28.73} & \textbf{38.70} & \textbf{53.09} \\ \hline
        \multirow{8}{*}{ViT} & \multirow{8}{*}{Trans} & \multicolumn{2}{l}{SeqMoCo} & 69.93 & 82.81 & 79.41 & 65.22 & 50.79 & 52.40 & 50.00 & 26.93 & 40.08 & 44.57 & 27.57 & 41.14 & 52.57 \\ 
        ~ & ~ & \multicolumn{2}{l}{SimMIM$^{\text{\textdagger}}$~\cite{simmim}} & 66.13 & 78.89 & 75.57 & 57.65 & 45.84 & 46.05 & 39.24 & 24.84 & 35.88 & 38.71 & 28.31 & 39.53 & 48.05 \\
        ~ & ~ & \multicolumn{2}{l}{iBOT$^{\text{\textdagger}}$~\cite{ibot}} & 64.53 & 80.85 & 75.47 & 64.45 & 45.59 & 46.67 & 43.40 & 20.48 & 32.82 & 39.30 & 23.88 & 36.75 & 47.85 \\
        ~ & ~ &  \multicolumn{2}{l}{DiG$^{\text{\textdagger}}$~\cite{dig}} & \underline{73.17} & \underline{86.04} & \underline{82.66} & \underline{68.62} & 52.62 & 55.81 & \underline{52.43} & \underline{29.67} & \underline{44.97} & \underline{46.93} & \underline{29.01} & \underline{42.51} & \underline{55.37} \\ 
        \cline{3-17}
        ~ & ~ & \multirow{3}{*}{RCLSTR} & w/ ERE & 70.73 & 85.47 & 79.90 & 65.53 & 50.89 & 53.49 & 48.26 & 25.98 & 38.80 & 44.57 & 28.06 & 41.80 & 52.79 \\
        ~ & ~ & & w/ ERE-INTRA & 71.90 & 84.31 & 80.79 & 67.23 & 51.76 & 54.57 & 51.74 & 27.17 & 42.75 & 45.52 & 28.19 & 40.89 & 53.90 \\
        ~ & ~ & & w/ ERE-INTRA-INTER & \underline{73.17} & 84.20 & \underline{82.66} & 68.47 & \underline{52.67} & \underline{59.22} & 50.00 & 28.49 & 42.81 & 45.62 & \textbf{30.05} & 42.26 & 54.97 \\ 
        ~ & ~ & \multicolumn{2}{l}{RCMSTR} & \textbf{75.80} & \textbf{87.66} & \textbf{85.32} & \textbf{73.26} & \textbf{56.67} & \textbf{61.71} & \textbf{59.03} & \textbf{31.59} & \textbf{48.28} & \textbf{50.39} & \textbf{30.05} & \textbf{44.54} & \textbf{58.69} \\ \hline
    \end{tabular}
\end{center}

\label{tab:representation_quality}

\end{table*}

\noindent\textbf{Settings.} 
Based on the evaluation of the decoder~\cite{aberdam2021sequence, liu2022perceiving, dig}, we freeze the base encoder during the training for feature representation evaluation. In this process, we only train the decoder to assess the quality of the feature representation. Considering the hardware constraints and the requirement of a large batch size for SeqCLR~\cite{aberdam2021sequence}, we choose MoCo~\cite{he2020momentum} as our baseline framework.
Building on MoCo's framework, we treat sequence features as atomic elements to implement a naive relational contrastive learning method (as in Equation~\ref{eq:relationloss}), which we refer to as SeqMoCo.
On the basis of SeqMoCo, we further explore the intra- and inter-hierarchical relations with the enriched textual information.
We add three modules that sequentially construct RCLSTR.
Then, we add the relational MIM module and construct the unified self-supervised framework named RCMSTR.
For CTC-based and attention-based decoders, we inherit the configurations from SeqCLR~\cite{aberdam2021sequence} and PerSec~\cite{liu2022perceiving}. For the Transformer-based decoder, we inherit the configurations from DiG~\cite{dig}. 

\noindent\textbf{Results.} 
We compare our results with other SSL methods for both CNN-based and ViT-based architectures, and the results are shown in Table~\ref{tab:representation_quality}. 
For the CNN-based results, RCMSTR achieves the best results on all benchmarks on both CTC-based and attention-based decoder configurations. Compared with SeqMoCo without KL loss, SeqMoCo with naive relational contrastive learning achieves limited performance gain. This performance gain is limited by finite dataset relations and suffers from over-fitting due to the lexical dependencies. Compared with SeqMoCo, RCLSTR is equipped with three modules of relational contrastive learning and further gains a significant improvement on average for the CTC-based and attention-based decoder. Also, the effectiveness of the three key modules is distinctly verified in this table. As we sequentially incorporate three modules, the performance consistently demonstrates improvement. Benefiting from our carefully crafted CNN-based MIM using sparse convolution, we construct a more powerful and unified RCMSTR. The unified RCMSTR further outperforms RCLSTR and achieves the best results on CTC and attention-based decoders.  

For the ViT-based results, we compare the RCMSTR approach with other ViT-based SSL text recognition methods in the bottom half of Table~\ref{tab:representation_quality}. We implement SeqMoCo and RCLSTR pre-training on the ViT encoder and evaluate their performance on the Transformer-based decoder. Based on the integration of RCLSTR and relational MIM, we construct the RCMSTR self-supervised framework. Compared with SeqMoCo, RCLSTR gains improvements on ViT architecture. Also, the effectiveness of three key modules on ViT is distinctly verified in this table. As we sequentially incorporate three modules, the performance improvement demonstrates their effectiveness for ViT architectures. Furthermore, the unified RCMSTR outperforms RCLSTR and achieves the best results on average performance. To fairly compare with DiG~\cite{dig}, we reproduce the DiG pre-training on the same datasets as in our work, denoted as DiG$^{\text{\textdagger}}$. Our RCMSTR significantly surpasses DiG$^{\text{\textdagger}}$ on average performance, which indicates that our relational MIM and decoupling framework is better than the naive MIM method of DiG.


\subsection{Fine-tuning  Evaluation.}

\begin{table*}
\caption{\textbf{Fine-tuning results.} Accuracy(\%) of fine-tuning the model with labeled data. We unfreeze the parameters of the pre-trained encoder and fine-tune the whole network. ``Supervised baseline" does not perform self-supervised pre-training, in which parameters are randomly initialized.}
\vspace{-0.3cm}
\begin{center}
    \centering
    \begin{tabular}{clccccccccccccc}
    \hline
         Arch. & Method & IIIT5K & IC03 & IC13 & SVT & IC15 & SVTP & CUTE & COCO & CTW & TT & HOST & WOST & Avg\\ \hline
        \multirow{5}{*}{CNN} & Supervised baseline & 84.40 & 91.81 & 89.16 & 83.62 & 68.05 & 73.33 & \underline{71.08} & 44.28 & 65.22 & 59.81 & 38.04 & 53.77 & 68.55 \\ 
        ~ & SeqCLR~\cite{aberdam2021sequence} & 82.90 & 92.20 & 87.90 & - & - & - & - & - & - & - & - & - & - \\ 
        ~ & PerSec-CNN~\cite{liu2022perceiving} & 84.20 & - & 88.90 & 82.40 & 68.20 & 73.60 & 68.40 & - & - & - & - & - & - \\ 
        ~ & SeqMoCo & 84.40 & \underline{92.73} & 89.85 & \underline{84.54} & \underline{69.30} & \underline{74.88} & 64.81 & \underline{45.63} & 64.18 & \underline{60.89} & 39.53 & 57.82 & 69.05\\ 
        ~ & RCLSTR & \underline{86.03} & \underline{92.73} & \textbf{91.13} & 83.15 & 69.15 & \underline{74.88} & 67.94 & 45.41 & \underline{66.12} & 60.65 & \underline{40.89} & \underline{59.89} & \underline{69.83} \\ 
        ~ & RCMSTR & \textbf{86.80} & \textbf{94.69} & \underline{89.95} & \textbf{86.55} & \textbf{75.03} & \textbf{78.91} & \textbf{75.96} & \textbf{50.44} & \textbf{68.36} & \textbf{65.73} & \textbf{47.85} & \textbf{65.11} & \textbf{73.78} \\
        \hline        
        \multirow{3}{*}{ViT} & Supervised baseline & 92.27 & 88.93 & 90.15 & 87.17 & 73.33 & 77.98 & 75.00 & 56.16 & 63.93 & 68.56 & 38.33 & 59.89 & 72.64 \\ 
        ~ & RCLSTR & \underline{95.57} & \underline{92.39} & \underline{93.50} & \underline{90.42} & \underline{79.59} & \textbf{85.89} & \underline{84.38} & \underline{64.26} & \underline{72.77} & \underline{76.19} & \underline{51.99} & \underline{68.42} & \underline{79.61} \\ 
        ~ & RCMSTR & \textbf{95.60} & \textbf{93.89} & \textbf{94.38} & \textbf{91.34} & \textbf{79.83} & \underline{85.58} & \textbf{88.54} & \textbf{64.30} & \textbf{73.73} & \textbf{77.51} & \textbf{53.64} & \textbf{70.86} & \textbf{80.77} \\ \hline
    \end{tabular}
\end{center}

\label{tab:finetune}

\end{table*}

\begin{table*}
\caption{\textbf{Semi-supervised results.} Accuracy(\%) of fine-tuning the model with 10\% and 1\% of the labeled data. We unfreeze the parameters of the pre-trained encoder and fine-tune the whole network. ``Supervised baseline" does not perform self-supervised pre-training, in which parameters are randomly initialized. The semi-supervised experiments use the CNN encoder and attention-based decoder here.}
\vspace{-0.3cm}
\begin{center}
    \centering
    \begin{tabular}{clccccccccccccc}
    \hline
        Fraction & Method & IIIT5K & IC03 & IC13 & SVT & IC15 & SVTP & CUTE & COCO & CTW & TT & HOST & WOST & Avg\\ \hline
        \multirow{3}{*}{10\%} & Supervised baseline & 70.90 & 83.85 & 79.01 & 66.46 & 49.74 & 50.70 & 47.04 & 32.64 & 44.78 & 44.24 & 20.03 & 32.41 & 51.82 \\ 
        ~ & SeqMoCo & 75.20 & \underline{87.77} & 81.87 & 71.41 & 54.47 & 57.98 & 48.78 & 34.88 & 52.09 & 49.04 & \underline{24.17} & 38.37 & 56.34 \\ 
        ~ & RCLSTR & \underline{76.80} & 87.31 & \underline{82.86} & \underline{72.64} & \underline{55.31} & \underline{60.16} & \underline{54.01} & \underline{36.54} & \underline{52.99} & \underline{49.60} & 23.30 & \underline{40.44} & \underline{57.66} \\ 
        ~ & RCMSTR & \textbf{84.13} & \textbf{93.08} & \textbf{89.66} & \textbf{83.31} & \textbf{69.25} & \textbf{74.73} & \textbf{69.34} & \textbf{45.57} & \textbf{65.90} & \textbf{61.54} & \textbf{41.06} & \textbf{57.08} & \textbf{69.55} \\
        \hline
        \multirow{3}{*}{1\%} & Supervised baseline & 64.57 & 80.05 & 74.09 & 60.59 & 42.92 & 45.12 & 37.28 & 26.87 & 37.46 & 39.16 & 16.68 & 28.10 & 46.07 \\ 
        ~ & SeqMoCo & 65.57 & 81.55 & 74.98 & 62.91 & 48.86 & 53.95 & 37.98 & 29.79 & 44.48 & 42.66 & 15.73 & 28.85 & 48.94  \\ 
        ~ & RCLSTR & \underline{73.73} & \underline{86.51} & \underline{81.77} & \underline{72.80} & \underline{51.35} & \underline{58.60} & \underline{45.99} & \underline{33.42} & \underline{50.90} & \underline{47.83} & \underline{18.75} & \underline{34.48} & \underline{54.68} \\ 
         ~ & RCMSTR & \textbf{79.90} & \textbf{89.16} & \textbf{85.12} & \textbf{79.91} & \textbf{61.19} & \textbf{64.96} & \textbf{62.37} & \textbf{39.17} & \textbf{59.33} & \textbf{54.27} & \textbf{28.44} & \textbf{44.37} & \textbf{62.35} \\ 
        \hline
    \end{tabular}

\end{center}

\label{tab:semi}
\end{table*}

\noindent\textbf{Settings.} During the training for the fine-tuning evaluation, the base encoder is not frozen, and we fine-tuned the whole network. Following~\cite{baek2019wrong},
we used ST~\cite{gupta2016synthetic} and MJ~\cite{jaderberg2014synthetic} as the fine-tuning training datasets. For CNN architecture, we employ the CNN backbone and attention-based decoder. For ViT architecture, we employ the ViT backbone and transformer-based decoder.

\noindent\textbf{Results.} 
Table~\ref{tab:finetune} shows the performance comparison between our proposed methods and others. ``Supervised baseline" does not perform self-supervised pre-training, in which parameters are randomly initialized. For CNN architectures, compared with SeqMoCo and supervised baseline, our RCLSTR method  gains an improvement on average performance. These results demonstrate that the image encoder learned by RCLSTR benefits downstream recognition fine-tuning. 
Moreover, compared with other counterparts (SeqCLR~\cite{aberdam2021sequence} and PerSec~\cite{liu2022perceiving}), the unified RCMSTR  significantly outperforms them on most datasets and average performance. The unified RCMSTR also outperforms RCLSTR and achieves the best results on average performance.  This indicates that integrating relational MIM and RCLSTR into the CNN downstream fine-tuning process can further enhance its performance.

For ViT architectures, a similar trend can be observed: our RCMSTR method outperforms both the randomly initialized supervised baseline and RCLSTR on average accuracy performance. These results show that the ViT encoder learned by relational MIM modeling is beneficial for downstream text recognition fine-tuning.

\subsection{Semi-Supervised Learning}

\noindent\textbf{Settings.} To assess downstream fine-tuning with limited amounts of data, we evaluate our method by considering semi-supervised settings. We use the same CNN encoders as before, which were pre-trained on the unlabeled data, and let the whole network be fine-tuned using 1\% or 10\% of the labeled dataset. We randomly select 1\% and 10\% of the dataset, ensuring that the same selected data is used across all experiments.

\noindent\textbf{Results.} Table~\ref{tab:semi} compares our method with other methods and supervised baseline training. It is evident that our method achieves the best performance on average across various amounts of labeled data. Our method succeeds in significantly improving the results of SeqMoCo. Compared with SeqMoCo, our RCLSTR gains a significant improvement on average for 10\% labeled data and 1\% labeled data. These results verify that the representations learned by RCLSTR enhance the learning from limited data. Furthermore, the unified RCMSTR outperforms RCLSTR and achieves the best results on both 1\% and 10\% labeled data. This demonstrates that the integration of relational MIM and RCLSTR also benefits the data-insuﬃcient tasks.




\subsection{Results on More Languages and Types of Text Image Datasets}

The use of our method is conditioned solely on the assumption that the text is horizontal, so it is potentially beneficial for other languages ($e.g.$, Chinese) and various fonts ($e.g.$, handwritten). 
Therefore, we experiment with different self-supervised pre-training methods on the Chinese document dataset~\cite{yu2021benchmarking}.
The settings remain consistent with those described in the representation quality evaluation, where we freeze the encoder to assess the decoder's performance. The accuracies obtained are summarized in Table~\ref{tab:handwritten}. . RCLSTR achieved better performance than SeqMoCo on Chinese datasets. Since Chinese Text Images have left-to-right structures and horizontal multi-grained hierarchies, RCLSTR can also facilitate self-supervised learning of their features. Furthermore, the unified RCMSTR outperforms RCLSTR and achieves the best results on average performance. This mainly comes from the fact that Chinese texts consist of radicals, and MIM can facilitate learning the local features. 

We further evaluate our RCLSTR and RCMSTR method on the handwritten datasets, comparing its performance with SeqMoCo. We consider the English handwritten datasets IAM~\cite{iam} and CVL~\cite{cvl}, and the results are summarized in the right part of Table~\ref{tab:handwritten}. Compared with SeqMoCo, RCLSTR achieves better performance on these two datasets and gains a large improvement for IAM and CVL. Although handwritten fonts exhibit a certain level of irregularity, our RCLSTR is capable of leveraging their horizontal and multi-hierarchical structure to enhance feature learning. Furthermore, the unified RCMSTR outperforms RCLSTR by a large margin for IAM and CVL. 

\subsection{Text Segmentation}

\begin{table}[t]
\caption{The decoder evaluation performance on Chinese and handwritten datasets. We trained a decoder with labeled data on top of frozen encoder, which was pre-trained on unlabeled images.}
\begin{center}
    \centering
    \begin{tabular}{ccccc}
    \hline
         \multirow{2}{*}{Method} & \multirow{2}{*}{Chinese Dataset} & \multicolumn{3}{c}{Handwritten Dataset}\\ 
          ~  & ~ & IAM & CVL & Avg \\ \hline 
          SeqMoCo & 47.56 & 56.16 & 77.80 & 66.98 \\
          RCLSTR & 55.70 & 62.88 & 85.92 & 74.40\\ 
          RCMSTR & \textbf{57.21} & \textbf{69.20} & \textbf{88.79} & \textbf{79.00} \\
          \hline
    
    \end{tabular}
\label{tab:handwritten}
\end{center}
\end{table}

\begin{table}[t]
\caption{Text segmentation fine-tune results on the TextSeg dataset. ``Scratch-ViT-Small" is trained from scratch, in which parameters are randomly initialized.}
\begin{center}
    \centering
    \begin{tabular}{cc}
    \hline
          Method & IoU \\ \hline 
          Scratch-ViT-Small & 81.1 \\
          RCLSTR-ViT-Small & 83.7 \\
          RCMSTR-ViT-Small & \textbf{83.9} \\ \hline
    
    \end{tabular}
\label{tab:seg}
\end{center}
\end{table}

\noindent\textbf{Settings.} We adopt the segmentation fine-tuning configuration from~\cite{dig} and use the TextSeg~\cite{textseg} dataset. TextSeg~\cite{textseg} is a finely annotated dataset designed for text segmentation tasks. Note that TextSeg is proposed for text segmentation on the whole image, while our task is performed on text patches. Therefore, we first obtain text patches using the bounding box annotations, ensuring that only one text instance is contained within each patch. During both training and evaluation phases, the text patches are resized to 32 × 128. The text segmentation model is composed of a ViT encoder and a text segmentation decoder. The decoder consists of 3 multi-head attention layers and a linear prediction head, with the number of heads set to 2 and the embedding dimension set to 384. An L1 loss is applied to the text segmentation model. 

\noindent\textbf{Results.} We conduct text segmentation experiments at the patch level on the TextSeg~\cite{textseg} dataset. Thus, we compare our text segmentation model, which includes RCMSTR pre-training, to the baseline that is trained from scratch. As shown in Table~\ref{tab:seg}, ``RCLSTR-ViT-Small" and ``RCMSTR-ViT-Small" outperform ``Scratch-ViT-Small" by 2.6\% and 2.8\% under the IoU metric, respectively. This indicates that text segmentation can benefit from relational contrastive learning and MIM pre-training.


% ...

\begin{table*}[htb]
\caption{\textbf{Ablations.} (a) Analysis of the setting for hierarchies. Without adding other modules, we experimented with various hierarchical combinations. (b) Effect of image division strategies. $^*$: Direct cutting is the default setting of RCLSTR. (c) Effect of inter-hierarchical combinations. (d) Ablation on inter-hierarchical loss functions.}
\vspace{-0.3cm}
\begin{center}
    \centering
    \begin{tabular}{clcccccccc}
    \hline
         \multirow{3}{*}{(a)} & Hierarchies  & IIIT5K & IC03 & IC13 & SVT & IC15 & SVTP & CUTE & Avg\\ \cline{2-10}        
         ~ & w/ subword \& frame  & 50.07 & 62.17 & 58.82 & 36.01 & 30.52 & 27.29 & 28.82 & 41.96\\ 
         ~ & w/ subword \& word   & 56.30& 66.90 & 64.63 & 43.59 & 35.19 & 35.50 & 35.07 & 48.17\\ 
         ~ & w/ subword \& word \& frame & \textbf{58.80} & \textbf{68.51} & \textbf{66.21} & \textbf{47.45} & \textbf{37.65} & \textbf{37.36} & \textbf{39.24} &\textbf{50.75}\\          
    \hline
         \multirow{3}{*}{(b)} & Image Division & ~ & ~ & ~ & ~ & ~ & ~ & ~ & ~ \\ \cline{2-10} 
         ~ & Direct cutting$^*$ & 61.07 & 72.90 & 68.77 & 50.54 & 40.30 & 40.16 & 39.24 & 53.28\\ 
         ~ & Dropping boundary features & 61.33 & 74.51 & 69.56 & 49.00 & 38.85 & 38.14 & 38.54 & 52.85 \\
         ~ & Vertical projection & 60.97 & 72.66 & 68.08 & 53.17 & 40.25 & 40.31 & 39.24 & 53.53 \\
    \hline
        \multirow{3}{*}{(c)} & Inter-Hierarchical Combinations  & ~ & ~ & ~ & ~ & ~ & ~ & ~ & ~\\ \cline{2-10}
         ~ & w/ subword-word  & 59.47 & 70.24 & 66.60 & 47.30 & 36.11 & 38.14 & 32.29 & 50.02\\ 
         ~ & w/ frame-subword  & 59.60 & 71.28 & 67.00 & 50.08 & 38.81 & \textbf{41.24} & 38.19 & 52.31\\ 
         ~ & w/ subword-word \& frame-subword  & \textbf{61.07} & \textbf{72.90} & \textbf{68.77} & \textbf{50.54} & \textbf{40.30} & 40.16 & \textbf{39.24} & \textbf{53.28}\\ 
    \hline
         \multirow{3}{*}{(d)} & \multicolumn{3}{l}{Inter-Hierarchical Loss Functions}  & ~ & ~ & ~ & ~ & ~ & ~\\ \cline{2-10}
         ~ & InfoNCE  & 60.60 & \textbf{73.82} & \textbf{70.15} & 48.38 & 39.38 & \textbf{40.16} & 37.15 & 52.81 \\ 
         ~ & KL  & \textbf{61.07} & 72.90 & 68.77 & \textbf{50.54} & \textbf{40.30} & \textbf{40.16} & \textbf{39.24} & \textbf{53.28}  \\ 
         ~ & RE  & 59.93 & 71.05 & 67.39 & 50.08 & 38.42 & 39.69 & 36.46 & 51.86 \\ \hline
    \end{tabular}

\end{center}
\vspace{-0.3cm}

\label{tab:ablation_good}
\end{table*}

         



\subsection{Visualization}

\begin{figure}[t]
\centering
\includegraphics[width=1\columnwidth]{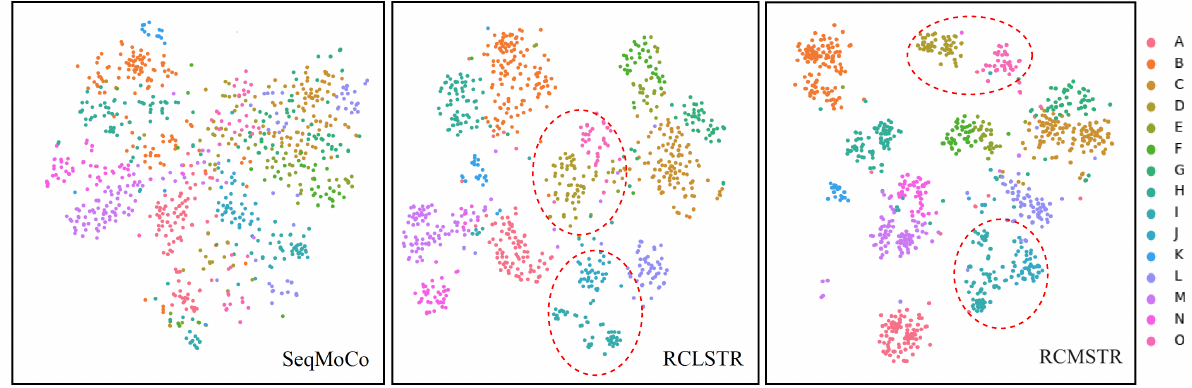}
\caption{t-SNE results.}
\label{fig:fig_tsne}
\end{figure}

In Figure~\ref{fig:fig_tsne}, we use t-SNE~\cite{tsne} to visualize the final features of IIIT5K~\cite{IIIT5K} images corresponding to SeqMoCo (baseline) and our proposed RCLSTR and RCMSTR, in which features for attention-based decoder are visualized by attaching
character labels to frame features. In our comparison between RCLSTR and SeqMoCo, RCLSTR enhances character clustering within the same class by mining intra-class relations. Besides, our RCLSTR also mines the inter-class relations, where clusters containing visually similar characters ($e.g.$, the circled I and J, as well as D and O) are positioned closely. On the other hand, RCMSTR and RCLSTR exhibit a comparable level of feature discrimination. Because the distinctiveness of representations is mainly from CL rather than MIM, the t-SNE of both is predominantly influenced by contrastive learning, resulting in similar feature discriminability. 

\subsection{Ablation Study}

         



\textbf{Effect of intra-hierarchy relations.}
Without adding other modules, we experimented with various hierarchical combinations, and the results are shown in Table~\ref{tab:ablation_good} (a). It is evident that the combination of subword and word levels outperforms the subword and frame levels in terms of performance. The highest performance is attained by learning across all three levels. In the following ablation, we used all three levels as the default setting to explore variants of other modules.

\begin{figure}[t]
  \centering
  \includegraphics[width=0.8\columnwidth]{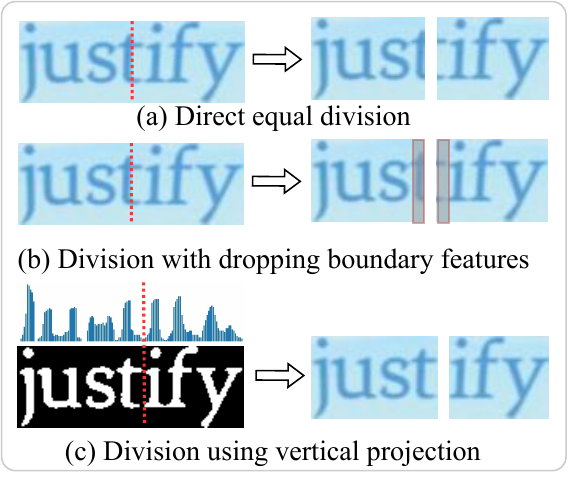}
  \caption{Multiple image division strategies. }
  \vspace{-0.5cm}
  \label{fig:imgdiv2}  
\end{figure}
\textbf{Effect of image division strategies.} We study how different image division strategies affect the effectiveness of relational regularization. The results under different image division strategies (as shown in Figure~\ref{fig:imgdiv2}) are summarized in Table~\ref{tab:ablation_good} (b). 
Under the condition of no available character positions for SSL, the process of direct division and concatenation creates richer context, but meaningless images (such as partial characters from non-ideal image division) are generated. There is a trade-off between context diversity and non-ideal image division. Since the KL loss can also serve as a regularization to avoid over-fitting the original context, the effectiveness mainly comes from more diversity of contexts, and it is insensitive to the non-ideal image division boundaries. We find that dropping the boundary features has a similar performance. The vertical projection method to avoid cutting character also has no significant performance gap. In supplementary materials, we provide samples of randomly permuted images.

\textbf{Effect of inter-hierarchy relations.} In Table~\ref{tab:ablation_good} (c), we impose subword-word and frame-subword consistency constraints separately. Our findings reveal that the frame-subword consistency yields the greatest improvement in performance. This indicates that the finer-grained consistency is more beneficial for learning text representations. 

\textbf{Ablation on inter-hierarchical loss functions.} By default, we use KL-divergence loss to constrain the consistency between hierarchies. As shown in Table~\ref{tab:ablation_good} (d), we test other consistency loss functions, such as InfoNCE loss in Equation~\ref{eq:infoloss} and relational loss (RE) in Equation~\ref{eq:relationloss}. In the context of cross-hierarchy relations, local and global features lack absolute consistency and exhibit only relative similarities. Consequently, KL loss proves to be more effective than both InfoNCE loss and the combination of RE (i.e., InfoNCE + KL) loss.

\begin{figure}[t]
\centering
\includegraphics[width=1\columnwidth]{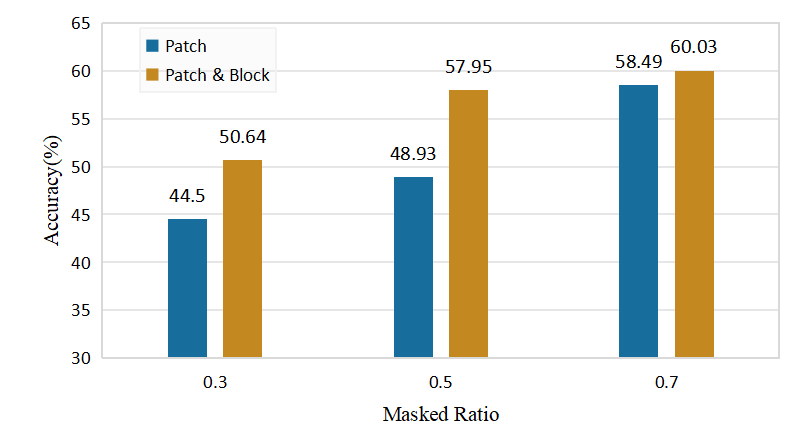}
\caption{Average evaluation of representations with different masked strategies and various masked ratios. Here, the accuracy denotes the average results on the first seven datasets (IIIT5K$ \sim $CUTE). The number of masked blocks is 1 here.}
\vspace{-0.5cm}
\label{fig:fig_ratio}
\end{figure}

\begin{figure}[t]
\centering
\includegraphics[width=1\columnwidth]{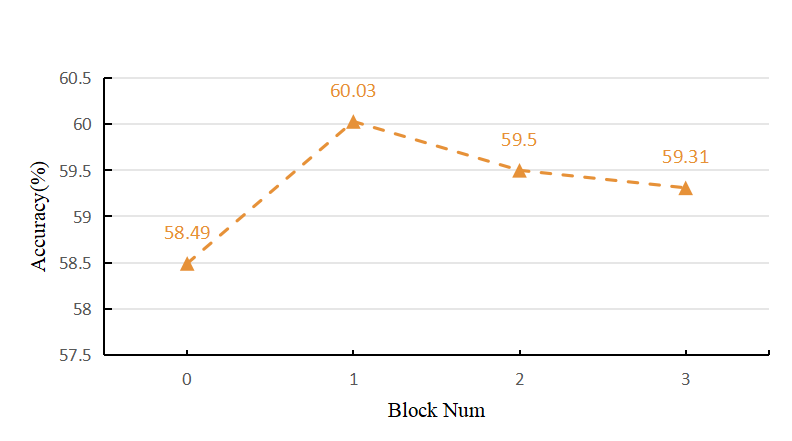}
\caption{Average evaluation of representations with various horizontal masked blocks. Here, the accuracy denotes the average results on the first seven datasets (IIIT5K$ \sim $CUTE). Block Num is the number of horizontal masked blocks applied on text images. }
\vspace{-0.5cm}
\label{fig:fig_blocknum}
\end{figure}

\textbf{Effect of masking strategy.} Our baseline method is a common randomly patch-masked MIM, which is denoted as the baseline method SimMIM~\cite{simmim}. Based on the random patch masking MIM, we add the horizontal block masking strategy. The patch masking masks local elements like strokes, while the block masking masks global elements like characters. Experimental results of transformer-based decoders are shown in Figure~\ref{fig:fig_ratio}. For various masked ratios, our patch and block joint masking always achieves higher performance than patch-only masking. For our default masked ratio of 0.7, the accuracy of transformer-based decoder is improved by an average of 1.54\% due to the use of both random patch masking and horizontal block masking strategies. This indicates that the local and global relational modeling of text images in MIM can achieve better representation quality.

\textbf{Number of horizontal masked blocks.} We applied different numbers of horizontal masked blocks for the ablation study. As shown in Figure~\ref{fig:fig_blocknum}, different amounts of horizontal blocks play the role of global relational modeling, so the performance is always better than normal random patch masking. The masking of 1 block achieves the highest performance improvement, and increasing the number of block masks has no further performance gain. This suggests that excessive masked blocks are sub-optimal, as they render the reconstruction task overly challenging and fail to further enhance self-supervised performance.

\section{Conclusions}
This work proposes a novel framework, Relational Contrastive Learning and Masked Image Modeling for Scene Text Recognition (RCMSTR). To take advantage of contextual priors in STR, we argue that contextual information can be interpreted as the relations of textual primitives and utilized in an unsupervised way. For relational contrastive learning, we propose RCLSTR with three modules to fully utilize the textual relations. 
With the permutation module, the textual relations are enriched on the fly, while the intra- and inter-hierarchy relational contrastive learning facilitates the capture of multi-granularity representations and interactions across different hierarchical levels, respectively.
Furthermore, we construct a unified RCMSTR framework for RCLSTR and relational MIM, and this framework further has three key contributions. First, we propose a joint reconstruction of masked patches and characters to fully learn about local and global relations. Second, our decoupled weight-sharing design effectively integrates the capabilities of both elements. Besides, our carefully designed framework integrates powerful MIM capabilities for CNN architectures. Extensive experiments verify the superiority of our RCMSTR method.

\bibliographystyle{IEEEtran}
\bibliography{reference.bib}

\begin{thebibliography}{10}
\providecommand{\url}[1]{#1}
\csname url@samestyle\endcsname
\providecommand{\newblock}{\relax}
\providecommand{\bibinfo}[2]{#2}
\providecommand{\BIBentrySTDinterwordspacing}{\spaceskip=0pt\relax}
\providecommand{\BIBentryALTinterwordstretchfactor}{4}
\providecommand{\BIBentryALTinterwordspacing}{\spaceskip=\fontdimen2\font plus
\BIBentryALTinterwordstretchfactor\fontdimen3\font minus \fontdimen4\font\relax}
\providecommand{\BIBforeignlanguage}[2]{{%
\expandafter\ifx\csname l@#1\endcsname\relax
\typeout{** WARNING: IEEEtran.bst: No hyphenation pattern has been}%
\typeout{** loaded for the language `#1'. Using the pattern for}%
\typeout{** the default language instead.}%
\else
\language=\csname l@#1\endcsname
\fi
#2}}
\providecommand{\BIBdecl}{\relax}
\BIBdecl

\bibitem{npid}
Z.~Wu, Y.~Xiong, S.~X. Yu, and D.~Lin, ``Unsupervised feature learning via non-parametric instance discrimination,'' in \emph{Proceedings of the IEEE conference on computer vision and pattern recognition}.\hskip 1em plus 0.5em minus 0.4em\relax IEEE, 2018, pp. 3733--3742.

\bibitem{cpc}
A.~v.~d. Oord, Y.~Li, and O.~Vinyals, ``Representation learning with contrastive predictive coding,'' \emph{arXiv preprint arXiv:1807.03748}, 2018.

\bibitem{he2020momentum}
K.~He, H.~Fan, Y.~Wu, S.~Xie, and R.~Girshick, ``Momentum contrast for unsupervised visual representation learning,'' in \emph{Proceedings of the IEEE/CVF conference on computer vision and pattern recognition}, 2020, pp. 9729--9738.

\bibitem{chen2020simple}
T.~Chen, S.~Kornblith, M.~Norouzi, and G.~Hinton, ``A simple framework for contrastive learning of visual representations,'' in \emph{International conference on machine learning}.\hskip 1em plus 0.5em minus 0.4em\relax PMLR, 2020, pp. 1597--1607.

\bibitem{beit}
H.~Bao, L.~Dong, S.~Piao, and F.~Wei, ``Beit: Bert pre-training of image transformers,'' \emph{arXiv preprint arXiv:2106.08254}, 2021.

\bibitem{mae}
K.~He, X.~Chen, S.~Xie, Y.~Li, P.~Doll{\'a}r, and R.~Girshick, ``Masked autoencoders are scalable vision learners,'' in \emph{Proceedings of the IEEE/CVF conference on computer vision and pattern recognition}, 2022, pp. 16\,000--16\,009.

\bibitem{simmim}
Z.~Xie, Z.~Zhang, Y.~Cao, Y.~Lin, J.~Bao, Z.~Yao, Q.~Dai, and H.~Hu, ``Simmim: A simple framework for masked image modeling,'' in \emph{Proceedings of the IEEE/CVF Conference on Computer Vision and Pattern Recognition}.\hskip 1em plus 0.5em minus 0.4em\relax IEEE COMPUTER SOC, 2022, pp. 9653--9663.

\bibitem{fang2021read}
S.~Fang, H.~Xie, Y.~Wang, Z.~Mao, and Y.~Zhang, ``Read like humans: Autonomous, bidirectional and iterative language modeling for scene text recognition,'' in \emph{Proceedings of the IEEE/CVF Conference on Computer Vision and Pattern Recognition}, 2021, pp. 7098--7107.

\bibitem{str2}
Y.~Wang, H.~Xie, S.~Fang, J.~Wang, S.~Zhu, and Y.~Zhang, ``From two to one: A new scene text recognizer with visual language modeling network,'' in \emph{Proceedings of the IEEE/CVF International Conference on Computer Vision}, 2021, pp. 14\,194--14\,203.

\bibitem{str3}
J.~Lee, S.~Park, J.~Baek, S.~J. Oh, S.~Kim, and H.~Lee, ``On recognizing texts of arbitrary shapes with 2d self-attention,'' in \emph{Proceedings of the IEEE/CVF Conference on Computer Vision and Pattern Recognition Workshops}, 2020, pp. 546--547.

\bibitem{aberdam2021sequence}
A.~Aberdam, R.~Litman, S.~Tsiper, O.~Anschel, R.~Slossberg, S.~Mazor, R.~Manmatha, and P.~Perona, ``Sequence-to-sequence contrastive learning for text recognition,'' in \emph{Proceedings of the IEEE/CVF Conference on Computer Vision and Pattern Recognition}, 2021, pp. 15\,302--15\,312.

\bibitem{liu2022perceiving}
H.~Liu, B.~Wang, Z.~Bao, M.~Xue, S.~Kang, D.~Jiang, Y.~Liu, and B.~Ren, ``Perceiving stroke-semantic context: Hierarchical contrastive learning for robust scene text recognition,'' in \emph{Proceedings of the AAAI Conference on Artificial Intelligence}, vol.~36, no.~2, 2022, pp. 1702--1710.

\bibitem{dig}
M.~Yang, M.~Liao, P.~Lu, J.~Wang, S.~Zhu, H.~Luo, Q.~Tian, and X.~Bai, ``Reading and writing: Discriminative and generative modeling for self-supervised text recognition,'' in \emph{Proceedings of the 30th ACM International Conference on Multimedia}, 2022, pp. 4214--4223.

\bibitem{vit}
A.~Dosovitskiy, L.~Beyer, A.~Kolesnikov, D.~Weissenborn, X.~Zhai, T.~Unterthiner, M.~Dehghani, M.~Minderer, G.~Heigold, S.~Gelly \emph{et~al.}, ``An image is worth 16x16 words: Transformers for image recognition at scale,'' \emph{arXiv preprint arXiv:2010.11929}, 2020.

\bibitem{maskocr}
P.~Lyu, C.~Zhang, S.~Liu, M.~Qiao, Y.~Xu, L.~Wu, K.~Yao, J.~Han, E.~Ding, and J.~Wang, ``Maskocr: Text recognition with masked encoder-decoder pretraining,'' \emph{arXiv preprint arXiv:2206.00311}, 2022.

\bibitem{context1}
D.~Yu, X.~Li, C.~Zhang, T.~Liu, J.~Han, J.~Liu, and E.~Ding, ``Towards accurate scene text recognition with semantic reasoning networks,'' in \emph{Proceedings of the IEEE/CVF conference on computer vision and pattern recognition}, 2020, pp. 12\,113--12\,122.

\bibitem{context2}
Z.~Qiao, Y.~Zhou, D.~Yang, Y.~Zhou, and W.~Wang, ``Seed: Semantics enhanced encoder-decoder framework for scene text recognition,'' in \emph{Proceedings of the IEEE/CVF conference on computer vision and pattern recognition}, 2020, pp. 13\,528--13\,537.

\bibitem{wan2020vocabulary}
Z.~Wan, J.~Zhang, L.~Zhang, J.~Luo, and C.~Yao, ``On vocabulary reliance in scene text recognition,'' in \emph{Proceedings of the IEEE/CVF Conference on Computer Vision and Pattern Recognition}, 2020, pp. 11\,425--11\,434.

\bibitem{zheng2021ressl}
M.~Zheng, S.~You, F.~Wang, C.~Qian, C.~Zhang, X.~Wang, and C.~Xu, ``Ressl: Relational self-supervised learning with weak augmentation,'' \emph{Advances in Neural Information Processing Systems}, vol.~34, pp. 2543--2555, 2021.

\bibitem{gupta2016synthetic}
A.~Gupta, A.~Vedaldi, and A.~Zisserman, ``Synthetic data for text localisation in natural images,'' in \emph{Proceedings of the IEEE conference on computer vision and pattern recognition}, 2016, pp. 2315--2324.

\bibitem{textseg}
X.~Xu, Z.~Zhang, Z.~Wang, B.~Price, Z.~Wang, and H.~Shi, ``Rethinking text segmentation: A novel dataset and a text-specific refinement approach,'' in \emph{Proceedings of the IEEE/CVF conference on computer vision and pattern recognition}, 2021, pp. 12\,045--12\,055.

\bibitem{rclstr}
J.~Zhang, T.~Lin, Y.~Xu, K.~Chen, and R.~Zhang, ``Relational contrastive learning for scene text recognition,'' in \emph{Proceedings of the 31st ACM International Conference on Multimedia}, 2023, pp. 5764--5775.

\bibitem{swav}
M.~Caron, I.~Misra, J.~Mairal, P.~Goyal, P.~Bojanowski, and A.~Joulin, ``Unsupervised learning of visual features by contrasting cluster assignments,'' \emph{Advances in neural information processing systems}, vol.~33, pp. 9912--9924, 2020.

\bibitem{byol}
J.-B. Grill, F.~Strub, F.~Altch{\'e}, C.~Tallec, P.~Richemond, E.~Buchatskaya, C.~Doersch, B.~Avila~Pires, Z.~Guo, M.~Gheshlaghi~Azar \emph{et~al.}, ``Bootstrap your own latent-a new approach to self-supervised learning,'' \emph{Advances in neural information processing systems}, vol.~33, pp. 21\,271--21\,284, 2020.

\bibitem{simsiam}
X.~Chen and K.~He, ``Exploring simple siamese representation learning,'' in \emph{Proceedings of the IEEE/CVF conference on computer vision and pattern recognition}, 2021, pp. 15\,750--15\,758.

\bibitem{dino}
M.~Caron, H.~Touvron, I.~Misra, H.~J{\'e}gou, J.~Mairal, P.~Bojanowski, and A.~Joulin, ``Emerging properties in self-supervised vision transformers,'' in \emph{Proceedings of the IEEE/CVF international conference on computer vision}, 2021, pp. 9650--9660.

\bibitem{ibot}
J.~Zhou, C.~Wei, H.~Wang, W.~Shen, C.~Xie, A.~Yuille, and T.~Kong, ``ibot: Image bert pre-training with online tokenizer,'' \emph{arXiv preprint arXiv:2111.07832}, 2021.

\bibitem{dinov2}
M.~Oquab, T.~Darcet, T.~Moutakanni, H.~Vo, M.~Szafraniec, V.~Khalidov, P.~Fernandez, D.~Haziza, F.~Massa, A.~El-Nouby \emph{et~al.}, ``Dinov2: Learning robust visual features without supervision,'' \emph{arXiv preprint arXiv:2304.07193}, 2023.

\bibitem{wang2022contrastive}
X.~Wang and G.-J. Qi, ``Contrastive learning with stronger augmentations,'' \emph{IEEE transactions on pattern analysis and machine intelligence}, vol.~45, no.~5, pp. 5549--5560, 2022.

\bibitem{whatif}
J.~Baek, Y.~Matsui, and K.~Aizawa, ``What if we only use real datasets for scene text recognition? toward scene text recognition with fewer labels,'' in \emph{Proceedings of the IEEE/CVF conference on computer vision and pattern recognition}, 2021, pp. 3113--3122.

\bibitem{siman}
C.~Luo, L.~Jin, and J.~Chen, ``Siman: Exploring self-supervised representation learning of scene text via similarity-aware normalization,'' in \emph{Proceedings of the IEEE/CVF Conference on Computer Vision and Pattern Recognition}, 2022, pp. 1039--1048.

\bibitem{guan2023self}
T.~Guan, W.~Shen, X.~Yang, Q.~Feng, Z.~Jiang, and X.~Yang, ``Self-supervised character-to-character distillation for text recognition,'' in \emph{Proceedings of the IEEE/CVF International Conference on Computer Vision}, 2023, pp. 19\,473--19\,484.

\bibitem{he2016reading}
P.~He, W.~Huang, Y.~Qiao, C.~Loy, and X.~Tang, ``Reading scene text in deep convolutional sequences,'' in \emph{Proceedings of the AAAI conference on artificial intelligence}, vol.~30, no.~1, 2016.

\bibitem{su2017accurate}
B.~Su and S.~Lu, ``Accurate recognition of words in scenes without character segmentation using recurrent neural network,'' \emph{Pattern Recognition}, vol.~63, pp. 397--405, 2017.

\bibitem{shi2016end}
B.~Shi, X.~Bai, and C.~Yao, ``An end-to-end trainable neural network for image-based sequence recognition and its application to scene text recognition,'' \emph{IEEE transactions on pattern analysis and machine intelligence}, vol.~39, no.~11, pp. 2298--2304, 2016.

\bibitem{shi2018aster}
B.~Shi, M.~Yang, X.~Wang, P.~Lyu, C.~Yao, and X.~Bai, ``Aster: An attentional scene text recognizer with flexible rectification,'' \emph{IEEE transactions on pattern analysis and machine intelligence}, vol.~41, no.~9, pp. 2035--2048, 2018.

\bibitem{yang2017learning}
X.~Yang, D.~He, Z.~Zhou, D.~Kifer, and C.~L. Giles, ``Learning to read irregular text with attention mechanisms.'' in \emph{IJCAI}, vol.~1, no.~2, 2017, p.~3.

\bibitem{wojna2017attention}
Z.~Wojna, A.~N. Gorban, D.-S. Lee, K.~Murphy, Q.~Yu, Y.~Li, and J.~Ibarz, ``Attention-based extraction of structured information from street view imagery,'' in \emph{2017 14th IAPR international conference on document analysis and recognition (ICDAR)}, vol.~1.\hskip 1em plus 0.5em minus 0.4em\relax Ieee, 2017, pp. 844--850.

\bibitem{lee2020recognizing}
J.~Lee, S.~Park, J.~Baek, S.~J. Oh, S.~Kim, and H.~Lee, ``On recognizing texts of arbitrary shapes with 2d self-attention,'' in \emph{Proceedings of the IEEE/CVF Conference on Computer Vision and Pattern Recognition Workshops}, 2020, pp. 546--547.

\bibitem{lu2021master}
N.~Lu, W.~Yu, X.~Qi, Y.~Chen, P.~Gong, R.~Xiao, and X.~Bai, ``Master: Multi-aspect non-local network for scene text recognition,'' \emph{Pattern Recognition}, vol. 117, p. 107980, 2021.

\bibitem{zhang2022context}
X.~Zhang, B.~Zhu, X.~Yao, Q.~Sun, R.~Li, and B.~Yu, ``Context-based contrastive learning for scene text recognition,'' in \emph{Proceedings of the AAAI Conference on Artificial Intelligence}, vol.~36, no.~3, 2022, pp. 3353--3361.

\bibitem{CLMIM}
N.~Park, W.~Kim, B.~Heo, T.~Kim, and S.~Yun, ``What do self-supervised vision transformers learn?'' \emph{arXiv preprint arXiv:2305.00729}, 2023.

\bibitem{maskedsiamese}
L.~Jing, J.~Zhu, and Y.~LeCun, ``Masked siamese convnets,'' \emph{arXiv preprint arXiv:2206.07700}, 2022.

\bibitem{convnextv2}
S.~Woo, S.~Debnath, R.~Hu, X.~Chen, Z.~Liu, I.~S. Kweon, and S.~Xie, ``Convnext v2: Co-designing and scaling convnets with masked autoencoders,'' in \emph{Proceedings of the IEEE/CVF Conference on Computer Vision and Pattern Recognition}, 2023, pp. 16\,133--16\,142.

\bibitem{IIIT5K}
A.~Mishra, K.~Alahari, and C.~Jawahar, ``Scene text recognition using higher order language priors,'' in \emph{BMVC-British machine vision conference}.\hskip 1em plus 0.5em minus 0.4em\relax BMVA, 2012.

\bibitem{ic03}
S.~M. Lucas, A.~Panaretos, L.~Sosa, A.~Tang, S.~Wong, R.~Young, K.~Ashida, H.~Nagai, M.~Okamoto, H.~Yamamoto \emph{et~al.}, ``Icdar 2003 robust reading competitions: entries, results, and future directions,'' \emph{International Journal of Document Analysis and Recognition (IJDAR)}, vol.~7, pp. 105--122, 2005.

\bibitem{ic13}
D.~Karatzas, F.~Shafait, S.~Uchida, M.~Iwamura, L.~G. i~Bigorda, S.~R. Mestre, J.~Mas, D.~F. Mota, J.~A. Almazan, and L.~P. De~Las~Heras, ``Icdar 2013 robust reading competition,'' in \emph{2013 12th international conference on document analysis and recognition}.\hskip 1em plus 0.5em minus 0.4em\relax IEEE, 2013, pp. 1484--1493.

\bibitem{svt}
K.~Wang, B.~Babenko, and S.~Belongie, ``End-to-end scene text recognition,'' in \emph{2011 International conference on computer vision}.\hskip 1em plus 0.5em minus 0.4em\relax IEEE, 2011, pp. 1457--1464.

\bibitem{ic15}
D.~Karatzas, L.~Gomez-Bigorda, A.~Nicolaou, S.~Ghosh, A.~Bagdanov, M.~Iwamura, J.~Matas, L.~Neumann, V.~R. Chandrasekhar, S.~Lu \emph{et~al.}, ``Icdar 2015 competition on robust reading,'' in \emph{2015 13th international conference on document analysis and recognition (ICDAR)}.\hskip 1em plus 0.5em minus 0.4em\relax IEEE, 2015, pp. 1156--1160.

\bibitem{svtp}
T.~Q. Phan, P.~Shivakumara, S.~Tian, and C.~L. Tan, ``Recognizing text with perspective distortion in natural scenes,'' in \emph{Proceedings of the IEEE international conference on computer vision}, 2013, pp. 569--576.

\bibitem{cute}
A.~Risnumawan, P.~Shivakumara, C.~S. Chan, and C.~L. Tan, ``A robust arbitrary text detection system for natural scene images,'' \emph{Expert Systems with Applications}, vol.~41, no.~18, pp. 8027--8048, 2014.

\bibitem{cocotext}
A.~Veit, T.~Matera, L.~Neumann, J.~Matas, and S.~Belongie, ``Coco-text: Dataset and benchmark for text detection and recognition in natural images,'' \emph{arXiv preprint arXiv:1601.07140}, 2016.

\bibitem{ctw}
Y.~Liu, L.~Jin, S.~Zhang, C.~Luo, and S.~Zhang, ``Curved scene text detection via transverse and longitudinal sequence connection,'' \emph{Pattern Recognition}, vol.~90, pp. 337--345, 2019.

\bibitem{tttext}
C.~K. Ch'ng and C.~S. Chan, ``Total-text: A comprehensive dataset for scene text detection and recognition,'' in \emph{2017 14th IAPR international conference on document analysis and recognition (ICDAR)}, vol.~1.\hskip 1em plus 0.5em minus 0.4em\relax IEEE, 2017, pp. 935--942.

\bibitem{ost}
Y.~Wang, H.~Xie, S.~Fang, J.~Wang, S.~Zhu, and Y.~Zhang, ``From two to one: A new scene text recognizer with visual language modeling network,'' in \emph{Proceedings of the IEEE/CVF International Conference on Computer Vision}, 2021, pp. 14\,194--14\,203.

\bibitem{shi2016robust}
B.~Shi, X.~Wang, P.~Lyu, C.~Yao, and X.~Bai, ``Robust scene text recognition with automatic rectification,'' in \emph{Proceedings of the IEEE conference on computer vision and pattern recognition}, 2016, pp. 4168--4176.

\bibitem{baek2019wrong}
J.~Baek, G.~Kim, J.~Lee, S.~Park, D.~Han, S.~Yun, S.~J. Oh, and H.~Lee, ``What is wrong with scene text recognition model comparisons? dataset and model analysis,'' in \emph{Proceedings of the IEEE/CVF international conference on computer vision}, 2019, pp. 4715--4723.

\bibitem{jaderberg2014synthetic}
M.~Jaderberg, K.~Simonyan, A.~Vedaldi, and A.~Zisserman, ``Synthetic data and artificial neural networks for natural scene text recognition,'' \emph{arXiv preprint arXiv:1406.2227}, 2014.

\bibitem{yu2021benchmarking}
H.~Yu, J.~Chen, B.~Li, J.~Ma, M.~Guan, X.~Xu, X.~Wang, S.~Qu, and X.~Xue, ``Benchmarking chinese text recognition: Datasets, baselines, and an empirical study,'' \emph{arXiv preprint arXiv:2112.15093}, 2021.

\bibitem{iam}
U.-V. Marti and H.~Bunke, ``The iam-database: an english sentence database for offline handwriting recognition,'' \emph{International Journal on Document Analysis and Recognition}, vol.~5, pp. 39--46, 2002.

\bibitem{cvl}
F.~Kleber, S.~Fiel, M.~Diem, and R.~Sablatnig, ``Cvl-database: An off-line database for writer retrieval, writer identification and word spotting,'' in \emph{2013 12th international conference on document analysis and recognition}, IEEE.\hskip 1em plus 0.5em minus 0.4em\relax IEEE, 2013, pp. 560--564.

\bibitem{tsne}
G.~E. Hinton and S.~Roweis, ``Stochastic neighbor embedding,'' \emph{Advances in neural information processing systems}, vol.~15, 2002.

\end{thebibliography}

\end{document}